\documentclass{article}

\PassOptionsToPackage{numbers, compress}{natbib}
\usepackage[preprint]{score_workshop}
\usepackage[utf8]{inputenc} % allow utf-8 input
\usepackage[T1]{fontenc}    % use 8-bit T1 fonts
\usepackage{xcolor}         % colors
\usepackage{hyperref}
\hypersetup{colorlinks=true,
            urlcolor=[RGB]{30, 144, 80},
            linkcolor=[RGB]{211, 64, 54},
            citecolor=[RGB]{5, 125, 238}}
\usepackage{url}            % simple URL typesetting
\usepackage{booktabs}       % professional-quality tables
\usepackage{amsfonts}       % blackboard math symbols
\usepackage{nicefrac}       % compact symbols for 1/2, etc.
\usepackage{microtype}      % microtypography

\usepackage{multirow}
\usepackage{subcaption}
\usepackage{amsmath,amssymb,amsfonts}
\usepackage{bm}
\usepackage{graphicx}
\usepackage{graphbox}

\usepackage{enumitem}
\setlist{nosep}

\usepackage{algorithm}
\usepackage{algorithmicx}
\usepackage[noend]{algpseudocode}
\usepackage{wrapfig}

\title{Conditional Diffusion Based on Discrete Graph Structures for Molecular Graph Generation}

\author{%
  Han Huang,\  Leilei Sun,\  Bowen Du,\  Weifeng Lv\\
  SKLSDE, Beihang University, Beijing, China\\
  \texttt{\{h-huang, leileisun, dubowen, lwf\}@buaa.edu.cn} \\
}

\newcommand{\ie}{\textit{i}.\textit{e}., }

\begin{document}

\maketitle

\begin{abstract}
Learning the underlying distribution of molecular graphs and generating high-fidelity samples is a fundamental research problem in drug discovery and material science. However, accurately modeling distribution and rapidly generating novel molecular graphs remain crucial and challenging goals.
To accomplish these goals, we propose a novel Conditional Diffusion model based on discrete Graph Structures (CDGS) for molecular graph generation. 
Specifically, we construct a forward graph diffusion process on both graph structures and inherent features through stochastic differential equations (SDE) 
and derive discrete graph structures as the condition for reverse generative processes.
We present a specialized hybrid graph noise prediction model that extracts the global context and the local node-edge dependency from intermediate graph states.
We further utilize ordinary differential equation (ODE) solvers for efficient graph sampling, based on the semi-linear structure of the probability flow ODE.
Experiments on diverse datasets validate the effectiveness of our framework.
Particularly, the proposed method still generates high-quality molecular graphs in a limited number of steps.
% Our code is provided in \href{https://github.com/GRAPH-0/CDGS}{https://github.com/GRAPH-0/CDGS}. 
\end{abstract}

\section{Introduction}

Dating back to the early works of Erd{\H{o}}s R{\'e}nyi random graphs \cite{erdHos1960evolution}, graph generation has been extensively studied for  applications in biology, chemistry, and social science.  
Recent graph generative models make great progress in graph distribution learning by exploiting the capacity of neural networks.
Models for molecular graph generation are notable for their success in representing molecule structures and restricting molecule search space, which facilitates drug discovery and material design.
In terms of the sampling process of graph generative models, autoregressive generation constructs molecular graphs step-by-step with decision sequences \cite{Youpolicy18, JinBJ18, Shigraphaf20, Luographdf21}, whereas one-shot generation builds all graph components at once \cite{zang2020moflow, lippe2020categorical, liu2021graphebm}.
Recently, diffusion-based models have been applied effectively to one-shot molecular graph generation \cite{JoLH22GDSS}, highlighting the advantages of flexible model architecture requirements and graph permutation-invariant distribution modeling. 

However, current diffusion-based models for molecular graphs still suffer from generation quality and sampling speed issues.
In \cite{JoLH22GDSS}, the generated graph distribution faces an obvious distance from the true distribution of datasets.
Furthermore, their sampling process relies heavily on extra Langevin correction steps \cite{song2021score} to diminish approximation errors, which largely increases computational cost and inference time, implying insufficient expressiveness of the graph score estimate model.
We argue that two major factors hinder the practice of diffusion-based models for molecular graph generation.
One is to focus on real-number graph formulation (\ie representing molecules as node feature and edge feature matrices) while neglecting the discrete graph structures, making it difficult to extract accurate local motifs from noisy real-number matrices for denoising and staying close to the true graph distribution.
The other is that a straightforward graph neural network design may not be strong enough to fully model the node-edge dependency from corrupted graphs and further satisfy the complex generation requirements, such as local chemical valency constraints, atom type proportion closeness, and global structure pattern similarity. 

To address these issues, we propose a novel Conditional Diffusion model based on discrete Graph Structures (CDGS) for molecular graph generation.
We find that considering graph discreteness and designing suitable graph noise prediction models could boost the ability of diffusion models in the graph domain, allowing for faster sampling and downstream applications.

\textit{Graph discreteness.} 
We develop a simple yet effective method for incorporating discrete graph structures without using special discrete state spaces.
Along with variables for node and edge features, additional one-bit discrete variables are added to indicate the potential existence of edges.
We convert them to real numbers and determine the quantization threshold.
In our diffusion framework, the continuous forward process is applied directly to edge existence variables, but for the reverse process, discrete graph structures are decoded first and serve as the condition for each sampling step.

\textit{Graph noise prediction model.}
We design a hybrid graph noise prediction model composed of standard message passing layers on discrete graphs and attention-based message passing layers on fully-connected graphs.
The first concentrates on neighbor node-edge dependency modeling, and the second on global information extraction and transmission.
Unlike \cite{JoLH22GDSS} which utilizes separate networks for node and edge denoising, we apply the unified graph noise prediction model to explicitly interact the node and edge representations from both real-valued matrices and discrete graph structures.

\textit{Fast sampling and downstream applications.}
We employ stochastic differential equations (SDEs) to describe the graph diffusion process.
With the simple Euler-Maruyama method, our diffusion-based model can obtain high-fidelity samples in $200$ steps of network evaluations, much fewer steps than the previous method.
We can benefit from recent research on probability flow ordinary differential equations (ODE) \cite{DEIS22, DPMS22} to promote fast graph sampling even further because we preserve the real-number graph description as an integral part of SDE.
Therefore, we introduce fast ODE solvers utilizing the semi-linear structure of probability flow ODEs for graphs.
Exploiting these ODE solvers, we also construct a useful pipeline for similarity-constrained molecule optimization based on latent space determined by the parameterized ODE and gradient guidance from the graph property predictor.

\begin{figure}[t]
    \centering
    
    \begin{subfigure}{0.49\textwidth}
    \centering
    \includegraphics[width=0.95\textwidth]{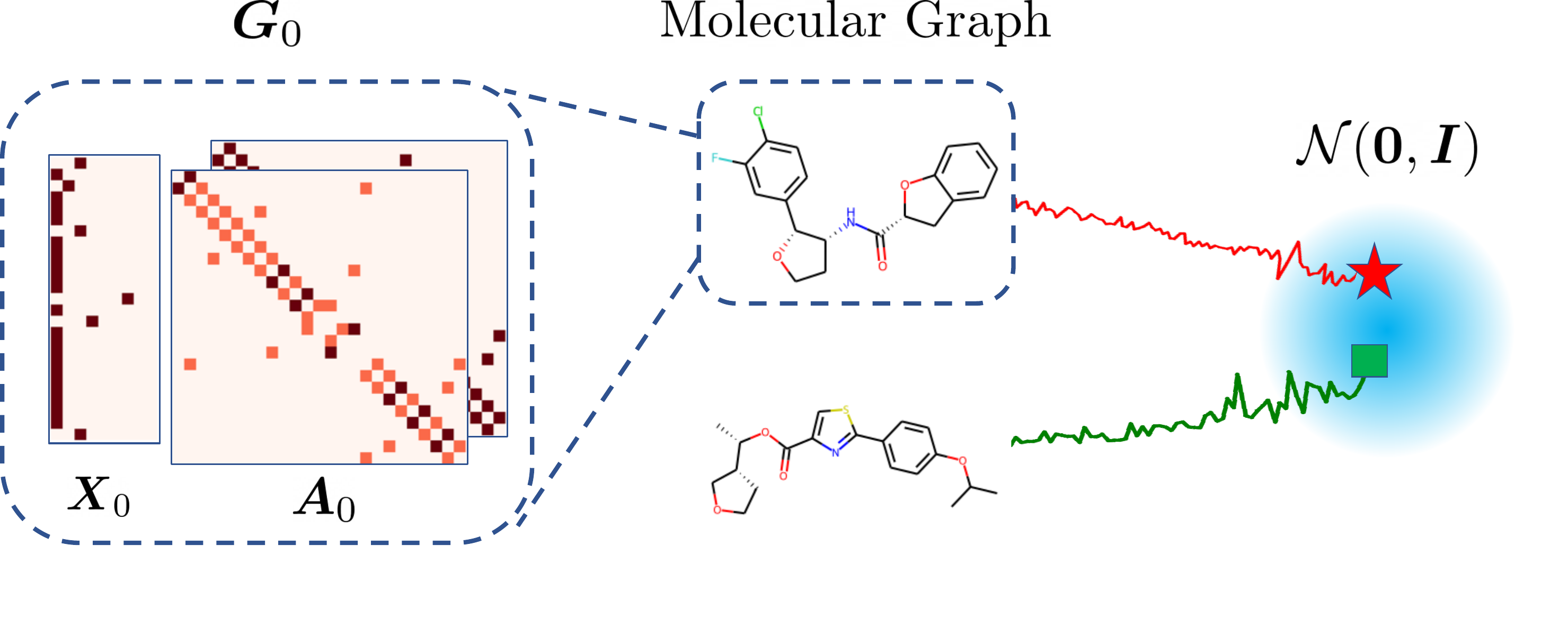}
    \end{subfigure}
    % \hfill
    \begin{subfigure}{0.49\textwidth}
    \centering
    \includegraphics[width=0.95\textwidth]{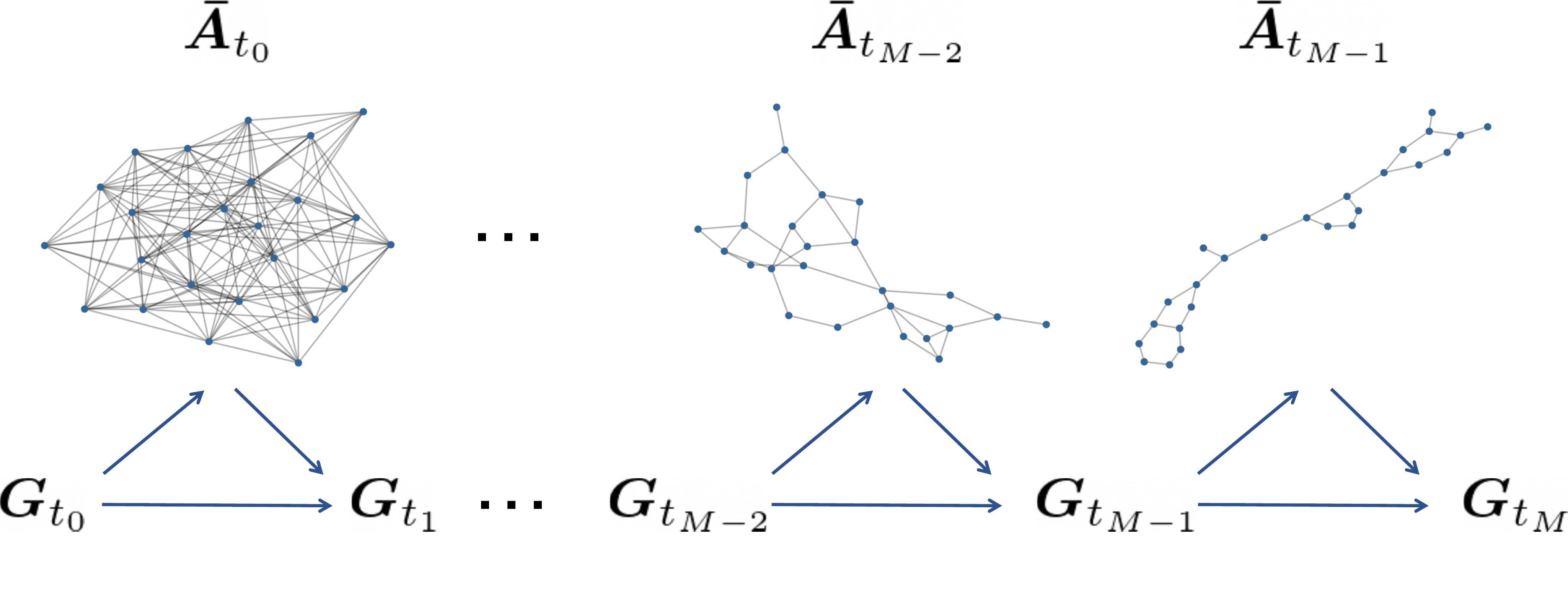}
    \end{subfigure}

    \caption{\textbf{(Left)} Forward diffusion process that perturbs molecular graphs towards a known prior distribution.
    A graph $\bm{G}_0$ is denoted by a node feature matrix $\bm{X}_0$ and a two-channel edge matrix $\bm{A}_0$ for edge types and existence.
    \textbf{(Right)} Discretized reverse generative process with discrete graph structure conditioning.}
    \label{fig:framework}
\end{figure}

Our main contributions are summarized as follows:
\vspace{-\topsep}
\begin{itemize}[leftmargin=*]
\item 
We propose a novel conditional diffusion framework based on discrete graph structures.
Leveraging a specialized graph noise prediction model, our framework accurately models the complex dependency between graph structures and features during the generative process.
\item 
We promote high-quality rapid graph sampling by adapting ODE solvers that utilize the semi-linear structure of the probability flow ODE. 
These ODE solvers also serve as the foundation for our effective similarity-constrained molecule optimization pipeline.
\item 
Experimental results demonstrate that our method outperforms the state-of-the-art baselines in both molecular graph and generic graph generation.
\end{itemize}

\section{Methodology}

\subsection{Conditional Graph Diffusion}
The first step in constructing diffusion probabilistic models \cite{sohl2015deep, ho2020denoising, song2021score, VDM2021} is to define a forward process that perturbs data with a sequence of noise until the output distribution becomes a known prior distribution. 
Assuming a continuous random variable $\bm{x}_0 \in \mathbb{R}^{d}$ and a well-defined forward process $\{\bm{x}_t\}_{t \in [0,T]}$,  we have
\begin{equation}
q_{0t}(\bm{x}_t|\bm{x}_0) = \mathcal{N}(\bm{x}_t | \alpha_t \bm{x}_0, {\sigma}^2_t \bm{I}) \ ,
\label{eq:q0t}
\end{equation}
where $\alpha_t, \sigma_t \in \mathbb{R}^{+} $ are time-dependant differentiable functions.
$\alpha_t$ and $\sigma_t$ are usually chosen to ensure that 
$q_T(\bm{x}_T) \approx \mathcal{N}(\bm{0}, \bm{I})$ with the decreasing signal-to-noise ratio $\alpha^2_t / \sigma^2_t$. 
By learning to reverse such a process, the diffusion model generates new samples from the prior distribution. 

It is a simple way to apply diffusion models to the graph domain by formulating graphs as high-dimensional variables $\bm{G} \in \mathbb{R}^{N \times F} \times \mathbb{R}^{N \times N}$ composed of $N$ node features with $F$ dimensions and an edge type matrix \cite{JoLH22GDSS}.
% In this formulation, the discrete graph structures are ignored throughout the process.
We argue that overlooked discrete graph structures, including motifs like rings and stars, may provide extra clues for node-edge dependency modeling and graph denoising.
We propose to separate the edge existence matrix from the edge type matrix and utilize a one-bit discrete variable representing the existence of a possible edge, forming ${\bar{\bm{A}}} \in \{0, 1\}^{N \times N}$ for the whole graph.
Instead of designing special discrete state spaces for discrete variables like \cite{HoogeboomArgmax21, AustinD3PM21}, we turn bits into real numbers and determine a quantization threshold.  %$\upsilon$.
Thus, we can conveniently apply continuous diffusion process to these variables and decode them with quantization back to discrete graph structure $\bar{\bm{A}_t}$ for $t \in [0,T]$.
The discrete graph structures can be plugged into the reverse process and function as conditions. 

% In addition to the corresponding $\bar{\bm{A}}$, 
We redefine the graph $\bm{G}$ by real-number node features $\bm{X} \in \mathbb{R}^{N \times F}$ and edge information $\bm{A} \in \mathbb{R}^{2 \times N \times N}$ (one channel for edge existence which can be quantized to $\bar{\bm{A}}$ and the other for edge types).  
The forward diffusion process for graphs shown in Figure \ref{fig:framework} can be described by the stochastic differential equation (SDE) sharing the same transition distribution in Eq. \ref{eq:q0t} \cite{VDM2021} with $t \in [0,T]$ as 
\begin{equation}
    \mathrm{d} \bm{G}_t = f(t) \bm{G}_t \mathrm{d} t + g(t) \mathrm{d} \bm{w}_t \ ,
    \label{eq:G_forward}
\end{equation}
where $f(t)=\frac{\mathrm{d} \log \alpha_t} {\mathrm{d}t} $ is the drift coefficient, 
$g^2(t)= \frac{\mathrm{d}\sigma^2_t}{\mathrm{d}t} - 2\frac{\mathrm{d} \log \alpha_t} {\mathrm{d}t}\sigma^2_t$ is the diffusion coefficient,
and $\bm{w}_t$ is a standard Wiener process.
The reverse-time SDE from time $T$ to $0$ \cite{song2021score} corresponding to Eq. \ref{eq:G_forward} is denoted as:
\begin{equation}
    \mathrm{d} \bm{G}_t = [f(t) \bm{G}_t - g^2(t) \nabla_{\bm{G}}\log q_t (\bm{G}_t)] \mathrm{d}_t 
    + g(t) \mathrm{d} \bar{\bm{w}}_t \ ,
\end{equation}
where $\nabla_{\bm{G}}\log q_t (\bm{G}_t)$ is the graph score function and $\bar{\bm{w}}_t$ is the reverse-time standard Wiener process. 
% To alleviate the expensive computation of high-dimensional graph scores,
We further split the reverse-time SDE into two parts that share the drift and diffusion coefficients as
\begin{equation}
    \left\{\begin{matrix}
    \mathrm{d} \bm{X}_t = [f(t) \bm{X}_t - g^2(t) \nabla_{\bm{X}}\log q_t (\bm{X}_t, \bm{A}_t)] \mathrm{d}_t + g(t) \mathrm{d} \bar{\bm{w}}^1_t
    \\
    \mathrm{d} \bm{A}_t = [f(t) \bm{A}_t - g^2(t) \nabla_{\bm{A}}\log q_t (\bm{X}_t, \bm{A}_t)] \mathrm{d}_t + g(t) \mathrm{d} \bar{\bm{w}}^2_t
    \end{matrix}\right.
.
\label{eq:graph_sde}
\end{equation}
We use a neural network $\bm{\epsilon_{\theta}}(\bm{G}_t, \bar{\bm{A}_t}, t)$ with discrete graph structure conditioning to parameterize the $\sigma_t$-scaled partial scores in Eq. \ref{eq:graph_sde}, where the node output of the neural network is denoted by $\bm{\epsilon}_{\bm{\theta}, \bm{X}}(\bm{G}_t, \bar{\bm{A}_t}, t)$ to estimate $-\sigma_t\nabla_{\bm{X}}\log q_t (\bm{X}_t, \bm{A}_t)$, and the edge output is denoted by $\bm{\epsilon}_{\bm{\theta}, \bm{A}}(\bm{G}_t, \bar{\bm{A}_t}, t)$ to estimate $-\sigma_t\nabla_{\bm{A}}\log q_t (\bm{X}_t, \bm{A}_t)$.
The model is optimized by the objective \cite{ho2020denoising, song2021score} as follows:
\begin{equation}
\begin{aligned}
    \min_{\bm{\theta}} \mathbb{E}_t\{ w(t) \mathbb{E}_{\bm{G}_0} \mathbb{E}_{\bm{G}_t|\bm{G}_0} 
    [ ||\bm{\epsilon}_{\bm{\theta}, \bm{X}}(\bm{G}_t, \bar{\bm{A}_t}, t) - \bm{\epsilon}_{\bm{X}}||^2_2 + 
    ||\bm{\epsilon}_{\bm{\theta}, \bm{A}}(\bm{G}_t, \bar{\bm{A}_t}, t) - \bm{\epsilon}_{\bm{A}}||^2_2]  \} \ ,
\end{aligned}
\end{equation}
where $w(t)$ is a given positive weighting function, $\bm{\epsilon}_{\bm{X}}$ and $\bm{\epsilon}_{\bm{A}}$ are the sampled Gaussian noise, and $\bm{G}_t = (\alpha_t \bm{X}_0 + \sigma_t \bm{\epsilon}_{\bm{X}}, \alpha_t \bm{A}_0 + \sigma_t \bm{\epsilon}_{\bm{A}})$.
In practice, we use the Variance-Preserving (VP) SDE for implementation, with the definition that $f(t)=-\frac{1}{2}\beta(t)$, $g(t)=\sqrt{\beta(t)}$, and $\beta(t) = \bar{\beta}_{min} + t(\bar{\beta}_{max} - \bar{\beta}_{min})$. 
With the optimized $\bm{\epsilon_{\theta}}$ and numerical solvers discretizing the SDE trajectory, shown in the right of Figure \ref{fig:framework}, new graph samples can be generated by solving the parameterized reverse-time SDE.

\subsection{Graph Noise Prediction Model}

Since $\bm{\epsilon_{\theta}}(\bm{G}_t, \bar{\bm{A}}_t, t)$ can be considered to predict the noise that is added to original graphs, we refer to it as the graph noise prediction model.
The design of noise prediction models plays a key role in diffusion-based generation, but it is still an open problem for the graph domain.
Applying the standard graph neural networks used in graph classification and link prediction tasks is not an appropriate choice due to the immediate real-number graph states and the complicated requirements for graph distribution learning.
In the case of molecular graphs, the model should focus on local node-edge dependence for chemical valency rules and attempt to recover global graph patterns like edge sparsity, frequent ring subgraphs, and even atom-type distribution.

To meet these challenges, we propose a hybrid message passing block (HMPB) consisting of two different kinds of message passing layers to explicitly model structure and feature dependency in both real-valued matrices ($\bm{X}_t$ and $\bm{A}_t$) and discrete graphs ($\bar{\bm{A}}_t$). 
One is a standard message passing layer like GINE \cite{GINE20} to aggregate local neighbor node-edge features, relying on the decoded discrete graph structures. 
The other one is a fully-connected attention-based message passing layer to focus on global information extraction and transmission.
We denote the node and edge representation update process in the $l$-th HMPB as
\begin{equation}
\begin{aligned}
    \bm{H}^{l+1}, \bm{E}^{l+1} & = \mathrm{HMPB}^{l}(\bm{H}^{l}, \bm{E}^{l}, \bar{\bm{A}}), \\
\mathrm{with} \ \ \ \ 
\bm{M}^{l+1} & = \mathrm{GINE}^l (\bm{H}^{l}, \bm{E}^{l}, \bar{\bm{A}}) + \mathrm{ATTN}^l (\bm{H}^{l}, \bm{E}^{l}), \\
\bm{H}^{l+1} & = \mathrm{FFN}_0^l(\bm{M}^{l+1}), \\
\bm{E}^{l+1}_{i,j} & = \mathrm{FFN}_1^l(\bm{M}^{l+1}_i + \bm{M}^{l+1}_j) ,
\end{aligned}
\end{equation}
where $\bm{H}^l \in \mathbb{R}^{N \times d}$ and $\bm{E}^l \in \mathbb{R}^{N \times N \times d}$ are node and edge inputs, $\bm{M}^{l+1} \in \mathbb{R}^{N \times d}$ is the aggregated message for nodes, $\bm{E}^{l+1}_{i,j} \in \mathbb{R}^{d}$ is the (i,j)-indexed edge output; $\mathrm{ATTN}^l$ is the full-connected attention layer; $\mathrm{FFN}^l$ is Feed Forward Network composed of the multilayer perceptron (MLP) and normalization layers. 
Here, the time $t$ and residual connections are omitted for clarity.
In particular, different from \cite{dwivedi2020generalization, graphformer21, kreuzer2021rethinking}, our attention layer takes edge features as the gate for both the message and dot-product calculation to thoroughly interact with node features and bias the message passing.
The key attention mechanism is denoted by
\begin{equation}
\small
    a_{i,j} = \mathrm{softmax}(\frac{(\mathrm{tanh}(\phi_0(\bm{E}_{i,j})) \cdot Q_i)K_j^\top}
    {\sqrt{d}}), \ 
    \mathrm{ATTN}_{i}(\bm{H}, \bm{E}) =  \sum_{j=0}^{N-1}a_{i,j} (\mathrm{tanh}(\phi_1(\bm{E}_{i,j})) \cdot V_j),
\end{equation}
where $Q, K, V$ are projected from node feature $\bm{H}$; $\bm{E}$ is the corresponding edge feature, $\phi_0$ and $\phi_1$ are learnable projections, and $\mathrm{tanh}$ is the activation layer.

For the initial features $\bm{H}^0$ and $\bm{E}^0$, we not only consider $\bm{X}_t$ and $\bm{A}_t$, but also extract structural encodings and relative positional encodings from $\bar{\bm{A}}_t$.
Using the $m$-step random walk matrix from the discrete adjacency matrix, we adopt the arrival probability vector as node features and obtain the truncated shortest-path distance from the same matrix as edge features. 
Time information is also added to the initial features with the sinusoidal position embedding \cite{vaswani2017attention}.
The final node and edge representations are respectively input to MLPs for graph noise prediction.
Note that without any node ordering dependent operations, our graph noise prediction model built upon message passing mechanisms is permutation equivariant and implicitly defines the permutation invariant graph log-likelihood function.

\subsection{ODE Solvers for Few-step Graph Sampling}
To generate graphs from the parameterized SDE in Eq. \ref{eq:graph_sde}, the SDE trajectory needs to be stimulated with numerical solvers.
The Euler-Maruyama (EM) solver is one of the simple and general solvers for SDEs.
Although our diffusion-based model can generate high-fidelity graphs in $200$ steps (a.k.a., number of function evaluation (NFE)) using the EM solver shown in Figure \ref{fig:fast sample},
such a solver still needs relatively long steps to achieve convergence in the high-dimensional data space and fails to meet the fast sampling requirement.
Since we preserve the continuous real-number graph diffusion formulation, one promising fast sampling method is to use the mature black-box ODE solvers for the probability flow ODE \cite{song2021score} that shares the same marginal distribution at time $t$ with the SDE.
Accordingly, the parameterized probability flow ODE for graphs is defined as  
\begin{equation}
    \mathrm{d} \bm{G}_t / \mathrm{d} t = f(t) \bm{G}_t + \frac{g^2(t)}{2\sigma_t} \bm{\epsilon_{\theta}}(\bm{G}_t, \bar{\bm{A}}_t, t) \ .
\label{eq:graph_ode}
\end{equation}
Recent works \cite{DEIS22, DPMS22} claim that the general black-box ODE solvers ignore the semi-linear structure of the probability flow ODE and introduce additional discretization errors. Therefore, new fast solvers are being developed to take advantage of the special structure of the probability flow ODE.

For our graph ODE in Eq. \ref{eq:graph_ode}, we further extend fast solvers based on the semi-linear ODE structure to generate high-quality graphs within a few steps.
By introducing $\lambda_t:= \log(\alpha_t/\sigma_t)$ and its inverse function $t_{\lambda}(\cdot)$ that satisfies $t=t_{\lambda}(\lambda(t))$, we change the subscript $t$ to $\lambda$ and get $\bm{\hat{G}}_\lambda:=\bm{G}_{t_\lambda(\lambda)}$,
$\bm{\hat{\epsilon}_\theta}(\bm{\hat{G}}_\lambda, {\bm{\bar{A}}}_\lambda', \lambda) := \bm{\epsilon_\theta}(\bm{G}_{t_\lambda(\lambda)}, {\bar{\bm{A}}_{t_\lambda(\lambda)}}, \lambda)$. 
%{\bar{\bm{A}}_\lambda}' = {\bar{\bm{A}}_{t_\lambda(\lambda)}},
We can derive the exact solution of the semi-linear probability flow ODE from time $s$ to time $t$ \cite{DPMS22} as 
\begin{equation}
\bm{G}_t = \frac{\alpha_t}{\alpha_s} \bm{G}_s - \alpha_t \int_{\lambda_s}^{\lambda_t}
e^{-\lambda} \bm{\hat{\epsilon}_\theta}(\bm{\hat{G}}_\lambda, {\bm{\bar{A}}}_\lambda', \lambda) \mathrm{d}{\lambda} \ .
\label{eq:exact_soluation}
\end{equation}
With the analytical linear part, we only need to approximate the exponentially weighted integral of $\bm{\hat{\epsilon}_\theta}$.
This approximation can be achieved by various methods \cite{HochbruckO05, HochbruckO10}, and we follow the derivation from \cite{DPMS22} to apply DPM-Solvers to graphs (denoted as GDPMS).
Given the initial graph sampled from the prior distribution $\bm{\tilde{G}}_{t_0} := \bm{G}_T=(\bm{X}_T, \bm{A}_T)$ with the predefined time step schedules $\{t_i\}_{i=0}^M$, the sequence $\{\bm{\tilde{G}}_{t_i} = \bm{(\tilde{X}}_{t_i}, \bm{\tilde{A}}_{t_i})\}_{i=1}^M$ is calculated iteratively by the first-order GDPMS as follows:
\begin{equation}
    \left\{\begin{matrix}
    \bm{\tilde{X}}_{t_i} = \frac{\alpha_{t_i}}{\alpha_{t_{i-1}}} \bm{\tilde{X}}_{t_{i-1}} - 
    \gamma_{i} \bm{\hat{\epsilon}_{\theta, X}}(\bm{\tilde{G}}_{t_{i-1}}, {\bm{\bar{A}}}_{t_{i-1}}', t_{i-1})
 \\
    \bm{\tilde{A}}_{t_i} = \frac{\alpha_{t_i}}{\alpha_{t_{i-1}}} \bm{\tilde{A}}_{t_{i-1}} - 
    \gamma_{i} \bm{\hat{\epsilon}_{\theta, A}}(\bm{\tilde{G}}_{t_{i-1}}, {\bm{\bar{A}}}_{t_{i-1}}', t_{i-1})
\end{matrix}\right. \ ,
\label{eq:GDPMS1}
\end{equation}
where $\gamma_{i} = \sigma_{t_i}(e^{\lambda_{t_i} - \lambda_{t_{i-1}}} - 1)$, and discrete graph structure ${\bm{\bar{A}}}_{t_{i-1}}'$ is decoded from $\bm{\tilde{G}}_{t_{i-1}}$. 
The final graph sample is derived from $\bm{\tilde{G}_{t_M}}$ with discretization. 
% More details about the sampling algorithms and high-order ODE  samplers for graphs are provided in Appendix.
More details on high-order ODE samplers for graphs are provided in Appendix.

\textbf{ODE-based Graph Optimization.}
Besides efficient sampling, the probability flow ODE offers latent representations for flexible data manipulation \cite{song2021score}.
Based on the latent space determined by the parameterized ODE and the graph DPM-Solvers assisted by gradient guidance, we propose a useful optimization pipeline for the meaningful similarity-constrained molecule optimization task.
% that mimics the drug discovery task of adjustment of an active starting molecule’s property while keeping similarity to the starting molecule to retain biological activity.

Specifically, we first train an extra time-dependent graph property predictor $\bm{R_\psi}(\bm{G}_t, t)$ on noisy graphs.
Then we setup a solver for the parameterized ODE in Eq. \ref{eq:graph_ode} to map the initial molecular graphs at time $0$ to the latent codes $\mathcal{G}_{t_\xi}$ at the time $t_\xi \in (0,T]$. 
Following the common optimization manipulation on latent space like \cite{JinBJ18,zang2020moflow}, we use the predictor to predict properties on the graph latent representation and lead the optimization towards molecules with desired properties through the gradient ascent, producing a latent graph sequence $\{{\bm{\mathcal{G}}^k_{t_\xi}}\}_{k=0}^K$.
% For the decoding process from the latent space to the molecular graph space, instead of using the same ODE as the forward encoding process, we introduce the gradient-guided ODE to further drive the sampling process to the high-property region.
Instead of using the same ODE as in the forward encoding process, we introduce the gradient-guided ODE to further drive the sampling process to the high-property region during the decoding process from the latent space to the molecular graph space.
The ODE with guidance can be modified from Eq. \ref{eq:graph_ode} as 
\begin{equation}
    \left\{\begin{matrix}
    \mathrm{d} \bm{X}_t / \mathrm{d} t = f(t) \bm{X}_t + \frac{g^2(t)}{2\sigma_t} 
    [\bm{\epsilon_{\theta, X}} - r \sigma_t \nabla_{\bm{X}}^{*}\bm{R_\psi}]
    \\
    \mathrm{d} \bm{A}_t / \mathrm{d} t = f(t) \bm{A}_t + \frac{g^2(t)}{2\sigma_t} 
    [\bm{\epsilon_{\theta, A}} - r \sigma_t \nabla_{\bm{A}}^{*}\bm{R_\psi}]
    \end{matrix}\right. \ ,
\end{equation}
where $r$ is the guidance weight, $\nabla^*$ refers to the unit normalized gradients, and the input $(\bm{G}_t, \bar{\bm{A}}_t, t)$ for $\bm{\epsilon_{\theta}}$ and $(\bm{G}_t, t)$ for $\bm{R_\psi}$ are omitted for simplicity. 
Notably, the GDPMS in Eq. \ref{eq:GDPMS1} can still work for the gradient-guided ODE by constructing the $\bm{\hat{\epsilon}_\theta}$ with the predictor gradients accordingly. 
The proposed pipeline can also be flexibly extended for multi-objective optimization by expanding the gradient guidance from multiple property prediction networks.

\section{Related Work}
\subsection{Molecule Generation}

Early attempts for molecule generation introduce sequence-based generative models and represent molecules as SMILES strings \cite{gomez2018automatic, KusnerPH17, DaiTDSS18}.
Besides the challenge from long dependency modeling, these methods may exhibit low validity rates since the SMILES string does not ensure absolute validity.
Therefore, graphs are more commonly used to represent molecule structures in recent studies.
Various graph generative models have been proposed to construct graphs autoregressively or in a one-shot form, based on different types of generative models, including variational auto-encoders \cite{simonovsky2018graphvae,LiuABG18}, generative adversarial networks \cite{de2018molgan, DEFactor18}, and normalizing flows \cite{Shigraphaf20, Luographdf21, lippe2020categorical, zang2020moflow}.
Compared to these models, our diffusion-based model advances in stable training and adaptable model architecture to consider the discrete graph structure for complicated dependency modeling. 
In addition, \cite{JinBJ18, ahn2022spanning} adopt an effective tree-based graph formulation for molecules, while our method keeps the general graph settings and models permutation invariant distributions.

\subsection{Diffusion Models} 

This new family of generative models \cite{sohl2015deep, ho2020denoising} correlated with score-based models \cite{song2021score, song2019generative} has demonstrated great power in the generation of high-dimensional data such as images.
For molecule science, in addition to molecular graph generation \cite{JoLH22GDSS}, diffusion models have also been applied to generate molecular conformations \cite{xu2022geodiff, torsional22} and 3D molecular structures \cite{HoogeboomEDM22}. 
Our framework greatly differs from the previous diffusion-based molecule generation in the conditional reverse process and the unified model design instead of separate models for nodes and edges.
Moreover, we promote efficient molecular graph generation with training-free samplers, which is primarily investigated in the image domain \cite{liu2022pseudo, DEIS22, DPMS22}.

\section{Experiment}
In this section, we display the experimental results of the proposed discrete graph structure assisted diffusion framework on multiple datasets. We provide more experiment details in Appendix.
Our code is provided in \href{https://github.com/GRAPH-0/CDGS}{https://github.com/GRAPH-0/CDGS}.
% and we will release the code in the future.
% [fast inference? conditional generation? In the form of question?]

\subsection{Molecular graph generation}

\subsubsection{Experimental Setup}
We train and evaluate models on two molecule datasets, ZINC250k \cite{zinc250k} and QM9 \cite{qm9}.
Before converting to graphs, all molecules are processed to the kekulized form using RDKit \cite{RDKit}, where hydrogen atoms are removed and aromatic bonds are replaced by double bonds. 
We evaluate generation quality on $10,000$ generated molecules with the following widely used metrics.
\textbf{Fr{\'{e}}chet ChemNet Distance (FCD)} \cite{FCD} calculates the distance between the reference molecule set and the generated set with the activations of the penultimate layer of ChemNet. Lower FCD values indicate higher similarity between the two distributions.
Following \cite{JoLH22GDSS}, we report FCD values after validity checking and valency correction since FCD is only calculated on valid molecules.
\textbf{Neighborhood subgraph pairwise distance kernel (NSPDK)} is the distance measured by mean maximum discrepancy (MMD), which incorporates node and edge features along with the underlying graph structure.
FCD and NSPDK, one from the perspective of molecules and the other from the perspective of graphs, are crucial for the evaluation of molecular graph distribution learning \cite{JoLH22GDSS}.
\textbf{VALID w/o check} is the percentage of valid molecules without post-hoc chemical valency correction.
Here, we follow the setting of \cite{zang2020moflow,JoLH22GDSS} to consider the formal charges for valency checking.
We also report the results of three metrics that are used commonly but have obvious marginal effects, \ie the ratio of valid molecules (\textbf{VALID}), the ratio of unique molecules (\textbf{UNIQUE}), and the ratio of novel molecules with reference to the training set (\textbf{NOVEL}). 

\begin{table*}[t]

\caption{Generation performance on ZINC250k \textbf{(Up)} and QM9 \textbf{(Down)}. The best results in first three metrics are highlighted in bold. The novelty metric on QM9 dataset denoted with $\star$ is debatable due to its contradiction with distribution learning.}
\label{tab:mol_res}

\begin{subfigure}{\textwidth}
\centering
\renewcommand\arraystretch{1.2}
\resizebox{0.9\textwidth}{!}{%
\begin{tabular}{cccccccc}
\hline
 &
  Method &
  \begin{tabular}[c]{@{}c@{}}VALID w/o\\ check (\%)\end{tabular} $\uparrow$ &
  NSPDK $\downarrow$ &
  FCD $\downarrow$ &
  VALID (\%) $\uparrow$ &
  UNIQUE (\%) $\uparrow$ &
  NOVEL (\%) $\uparrow$ \\ \hline
                           & \textit{Train}& -     & \textit{5.91e-5} & \textit{0.985} & - & - & - \\
                           \hline
\multirow{4}{*}{Autoreg.}  & GraphAF       & 68.00 & 0.044 & 16.289 & 100.00 & 99.10 & 100.00 \\
                           & GraphAF+FC    & 68.47 & 0.044 & 16.023 & 100.00 & 98.64 & 99.99  \\
                           & GraphDF       & 89.03 & 0.176 & 34.202 & 100.00 & 99.16 & 100.00 \\
                           & GraphDF+FC    & 90.61 & 0.177 & 33.546 & 100.00 & 99.63 & 100.00 \\ \hline
\multirow{11}{*}{One-shot} & MoFlow        & 63.11 & 0.046 & 20.931 & 100.00 & 99.99 & 100.00 \\
                           & GraphCNF      & 96.35 & 0.021 & 13.532 & 100.00 & 99.98 & 100.00 \\
                           & EDP-GNN       & 82.97 & 0.049 & 16.737 & 100.00 & 99.79 & 100.00 \\
                           & GraphEBM      & 5.29  & 0.212 & 35.471 & 99.96  & 98.79 & 100.00 \\ \cline{2-8} 
                           & GDSS          & 97.01 & 0.019 & 14.656 & 100.00 & 99.64 & 100.00 \\
                           & GDSS-EM       & 15.97 & 0.075 & 24.310 & 100.00 & 100.00 & 100.00 \\
                           & GDSS-VP-EM    & 33.01 & 0.048 & 24.471 & 100.00 & 100.00 & 100.00 \\ \cline{2-8} 
                           & CDGS-EM       & \textbf{98.13} & \textbf{7.03e-4} & \textbf{2.069}  & 100.00 & 99.99 & 99.99  \\
                           & CDGS-GDPMS-200 & 96.19 & 0.001 & 3.037  & 100.00 & 99.98 & 99.99  \\
                           & CDGS-GDPMS-50  & 95.56 & 0.002 & 3.567 & 100.00 & 99.98 & 99.99  \\
                           & CDGS-GDPMS-30  & 93.49 & 0.003 & 4.498 & 100.00 & 99.99 & 99.99  \\ \hline
\end{tabular}%
}
\end{subfigure}

\quad

\quad

\begin{subfigure}{\textwidth}
\centering
\renewcommand\arraystretch{1.1}
\resizebox{0.9\textwidth}{!}{%
\begin{tabular}{cccccccc}
\hline
 &
  Method &
  \begin{tabular}[c]{@{}c@{}}VALID w/o\\ check (\%) \end{tabular} $\uparrow$ &
  NSPDK $\downarrow$ &
  FCD $\downarrow$ &
  VALID (\%) $\uparrow$ &
  UNIQUE (\%) $\uparrow$ &
 \textit{ NOVEL (\%) } $\star$\\ \hline
                           & \textit{Train}& -     & \textit{1.36e-4} & \textit{0.057} & -      & -     & -      \\
                           \hline
\multirow{4}{*}{Autoreg.}  & GraphAF       & 67.00 & 0.020 & 5.268  & 100.00 & 94.51 & 88.83 \\
                           & GraphAF+FC    & 74.43 & 0.021 & 5.625  & 100.00 & 88.64 & 86.59 \\
                           & GraphDF       & 82.67 & 0.063 & 10.816 & 100.00 & 97.62 & 98.10 \\
                           & GraphDF+FC    & 93.88 & 0.064 & 10.928 & 100.00 & 98.58 & 98.54 \\ \hline
\multirow{10}{*}{One-shot} & MoFlow        & 91.36 & 0.017 & 4.467  & 100.00 & 98.65 & 94.72 \\
                           & EDP-GNN       & 47.52 & 0.005 & 2.680  & 100.00 & 99.25 & 86.58 \\
                           & GraphEBM      & 8.22  & 0.030 & 6.143  & 100.00 & 97.90 & 97.01 \\ \cline{2-8} 
                           & GDSS          & 95.72 & 0.003 & 2.900  & 100.00 & 98.46 & 86.27 \\
                           & GDSS-EM       & 66.01 & 0.016 & 5.112  & 100.00 & 90.05 & 94.24 \\
                           & GDSS-VP-EM    & 86.02 & 0.013 & 4.588  & 100.00 & 89.03 & 88.63 \\ \cline{2-8} 
                           & CDGS-EM       & \textbf{99.68} & \textbf{3.08e-4} & \textbf{0.200}  & 100.00 & 96.83 & 69.62 \\
                           & CDGS-GDPMS-200 & 99.54 & 3.68e-4 & 0.269  & 100.00 & 97.20 & 72.52 \\
                           & CDGS-GDPMS-50  & 99.47 & 3.85e-4 & 0.289  & 100.00 & 97.27 & 72.38 \\
                           & CDGS-GDPMS-30  & 99.18 & 4.13e-4 & 0.326  & 100.00 & 97.42 & 72.52 \\ \hline
\end{tabular}%
}    
\end{subfigure}
\end{table*}

\subsubsection{Baselines}
We compare our CDGS with several autoregressive and one-shot molecular graph generative models, including \textbf{GraphAF} \cite{Shigraphaf20}, \textbf{GraphDF} \cite{Luographdf21}, \textbf{MoFlow} \cite{zang2020moflow},  \textbf{GraphCNF} \cite{lippe2020categorical}, \textbf{EDP-GNN} \cite{niu2020permutation}, 
\textbf{GraphEBM} \cite{liu2021graphebm}, and \textbf{GDSS} \cite{JoLH22GDSS}.
\textbf{GraphAF+FC} and \textbf{GraphDF+FC} are the modified versions considering formal charges for fair comparison.
\textbf{GDSS-EM} is the result sampled with the EM solver, and \textbf{GDSS-VP-EM} is retrained with VPSDE, sharing the same SDE parameters with our model.

\subsubsection{Generation Quality}
The molecular graph generation quality benchmark results on ZINC250k and QM9 are reported in Table \ref{tab:mol_res}.
We run three times for our method and report the mean performance.
We provide the performance bound on two distribution metrics by measuring the distance between preprocessed training molecules and original test molecules. 
In the first three non-trivial metrics across two different molecule datasets, CDGS with the EM solver outperforms state-of-the-art molecular graph generative models.
The high validity rate before valency checking shows that CDGS learns the chemical valency rule successfully and avoids unrealistically frequent valency correction.
Furthermore, with much lower NSPDK and FCD values, CDGS learns the underlying distribution more faithfully in both graph and chemical space.
CDGS achieves such performance without any Langevin correction steps in sampling, while previous diffusion-based GDSS drops off obviously with the pure EM solver.
Using the same SDE parameters, the performance gap between GDSS-VP-EM and CDGS-EM further demonstrates the effectiveness of our framework design.
Another noteworthy point is that, equipped with the 3rd-order GDPMS, our proposed model maintains excellent generation ability with limited NFE decreasing from $200$ to $30$. 
Extra visualization of generated molecules is provided in Appendix. 

We also point out that the novelty metric on the QM9 dataset seems debatable because the QM9 dataset is almost an exhaustive list of molecules that adhere to a predetermined set of requirements \cite{vignac2022topn, HoogeboomEDM22}. Therefore, a molecule that is thought to be novel violates the constraints, which means the model is unable to capture the dataset properties. This metric is kept for experiment completeness.

\subsubsection{Fast Sampling}
\begin{figure}[t]
\centering
\begin{minipage}[t]{0.5\textwidth}
    \centering
    \begin{subfigure}{\textwidth}
        \includegraphics[width=\textwidth, align=t]{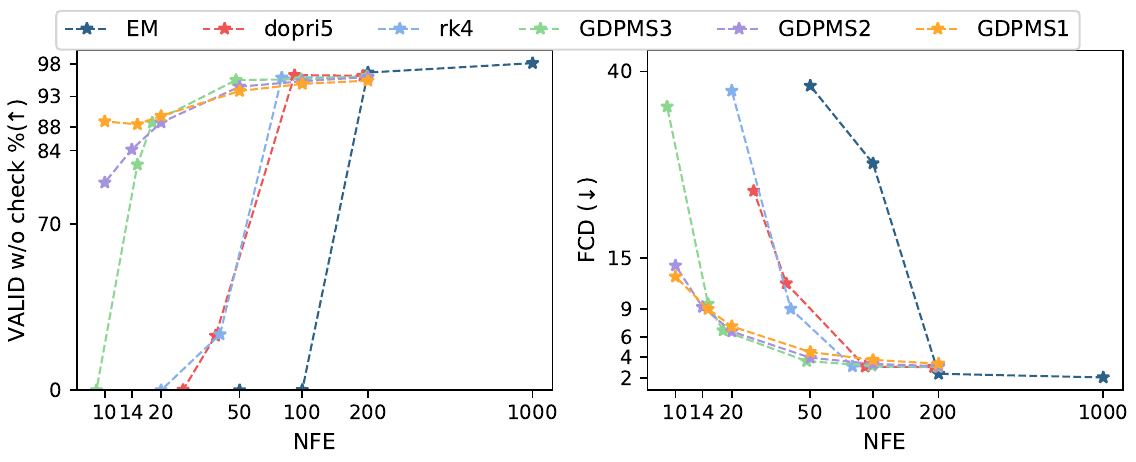}
    \end{subfigure}
    \begin{subfigure}{\textwidth}
        \renewcommand\arraystretch{2.0}
        \resizebox{\textwidth}{!}{
        \begin{tabular}{c|ccccccc}
        \hline
        Method   & GraphAF   & GraphDF   & MoFlow & GDSS      & CDGS-50   & CDGS-30   & CDGS-10 \\ \hline
        Time (s) & 2.89$e^2$ & 3.19$e^3$ & 1.54   & 1.42$e^2$ & 3.79$e^1$ & 2.38$e^1$ & 8.38    \\ \hline
    \end{tabular}}
    \end{subfigure}
    \caption{(Up) Molecular graph sampling results for various numerical solvers. (Down) The wall-clock time taken to generate $512$ molecular graphs.}
    \label{fig:fast sample}
\end{minipage}
\begin{minipage}[t]{0.47\textwidth}
    \centering
    \includegraphics[width=\textwidth, align=t]{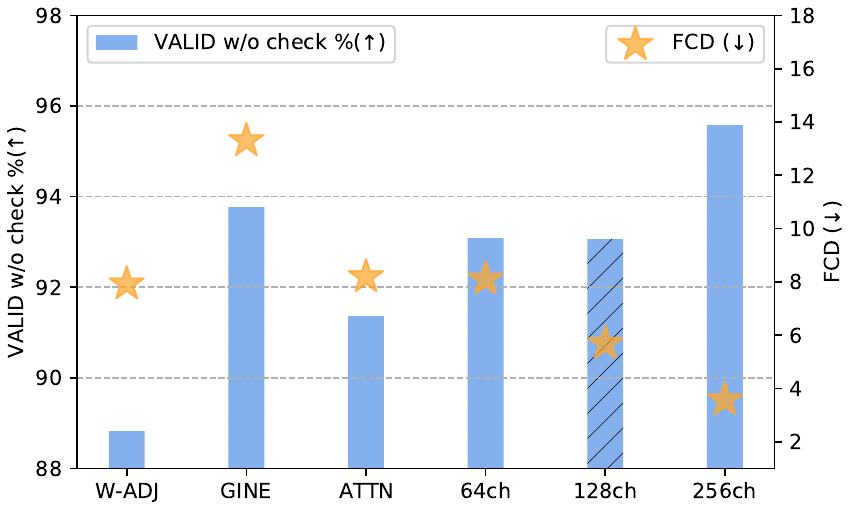}
    \caption{Ablation studies on ZINC250k.}
    \label{fig:abl}
\end{minipage}
\end{figure}

% \begin{wrapfigure}{r}{0.5\textwidth}
%     \centering
%     \begin{subfigure}{0.5\textwidth}
%         \centering
%         \includegraphics[width=\textwidth]{figs/fast_sampling.pdf}
%     \end{subfigure}
%     \begin{subfigure}{0.5\textwidth}
%         \centering
%         \renewcommand\arraystretch{2.0}
%         \resizebox{\textwidth}{!}{%
%         \begin{tabular}{c|ccccccc}
%         \hline
%         Method   & GraphAF   & GraphDF   & MoFlow & GDSS      & CDGS-50   & CDGS-30   & CDGS-10 \\ \hline
%         Time (s) & 2.89$e^2$ & 3.19$e^3$ & 1.54   & 1.42$e^2$ & 3.79$e^1$ & 2.38$e^1$ & 8.38    \\ \hline
%     \end{tabular}}
%     \end{subfigure}
%     \caption{(Up) Few-step molecular graph sampling results for various numerical solvers. (Down) The wall-clock time taken to generate $512$ molecular graphs.}
%     \label{fig:fast sample}
% \end{wrapfigure}
To explore fast and high-quality few-step molecular graph sampling, we compare the sampling quality of CDGS with different types of numerical solvers, including GDPMS with different orders, the EM solver, and black-box ODE solvers.  
For black-box ODE solvers, we pick out an adaptive-step and a fixed-step neural ODE solver implemented by \cite{chen2018neuralode}, that is, Runge-Kutta of order 5 of Dormand-Prince-Shampine (dopri5) and Fourth-order Runge-Kutta with 3/8 rule (rk4).
As shown in Figure \ref{fig:fast sample}, based on our conditional diffusion framework, the EM solver generates high-quality graphs between $200$ NFE and $1000$ NFE, but fails to converge under fewer NFE.
The black-box neural ODE solvers can obtain acceptable quality at around $50$ NFE.
The GDPMS displays clear superiority in the range below $50$ NFE.
Notably, the 1st-order GDPMS still generates reasonable molecular graphs with $10$ NFE.
For the running time comparison, CDGS equipped with GDPMS takes much less time compared to autoregressive GraphAF and GraphDF, and makes an obvious improvement towards GDSS.
MoFlow spends the least time but fails to generate high-fidelity samples according to Table \ref{tab:mol_res}.
In conclusion, benefiting from the framework design and the ODE solvers utilizing the semi-linear structure, we achieve great advancement in fast sampling for complex molecular graphs.

% \begin{wrapfigure}{r}{0.4\textwidth}
%     \centering
%     \includegraphics[width=0.4\textwidth]{figs/abl_study.pdf}
%     \caption{Ablation on ZINC250k.}
%     \label{fig:abl}
% \end{wrapfigure}

\subsubsection{Ablation Studies}

We conduct ablation analysis on the ZINC250k dataset to verify the effectiveness of our framework.
In Figure \ref{fig:abl}, with the goal to generate high-quality molecular graphs efficiently, we report the results using GDPMS with $50$ NFE, which is sufficient to obtain converged samples.
% also alleviating the disturbance from the stochastic term in the EM solver and faithfully reflecting the model ability.
Taking CDGS with $64$ hidden dimensions (\textbf{64ch}) as reference, we first remove the discrete graph structure related components and remain with our edge-gated attention layers (\textbf{ATTN}), then further remove the edge existence variable (\textbf{W-ADJ}). The variant using GINE without attention layers is denoted as \textbf{GINE}. 

We emphasize that VALID w/o check and FCD metrics are complementary and should be combined to assess molecule generation quality, because the former only reflects the valency validity of local atom and bond connections, whereas the latter is obtained after valency corrections and focuses more on global molecule similarity.
It can be observed from Figure \ref{fig:abl} that:
(1) Compared to 64ch, ATTN has a lower validity rate and gets a close FCD after more undesirable corrections, while GINE achieves high validity rates but fails to capture more global information.
It proves that the proposed attention module is crucial for global distribution learning and that discrete graph structures greatly help to capture the chemical valency rule.
(2) The comparison of W-ADJ and ATTN shows that separating the edge existence in the formulation also makes contributions to molecule validity.
In addition, W-ADJ outperforms GDSS-VP-EM in Table \ref{tab:mol_res}, showing the effectiveness of explicitly interacting node and edge representations using a unified graph noise prediction model.
(3) It is necessary to increase hidden dimensions (\textbf{128ch}, \textbf{256ch}) to better handle the complexity of drug-like molecules in the ZINC250k dataset.

\begin{table}[t]
\centering
\caption{Similarity-constrained molecule property optimization performance. The values above and below arrows in visualizations denote similarity scores and improvements.}
\label{tab:const_opt}

\begin{subfigure}[c]{0.49\textwidth}
\centering
\renewcommand\arraystretch{1.15}
\resizebox{\textwidth}{!}{%
\begin{tabular}{ccccc}
\hline
                     & \multicolumn{2}{c}{GraphAF-RL}          & \multicolumn{2}{c}{MoFlow}              \\
$\delta$             & \textbf{Improvement} & \textbf{Success} & \textbf{Improvement} & \textbf{Success} \\ \hline
0.0                  & 13.13$\pm$6.89       & 100\%            & 8.61$\pm$5.44        & 99\%             \\
0.2                  & 11.90$\pm$6.86       & 100\%            & 7.06$\pm$5.04        & 97\%             \\
0.4                  & 8.21$\pm$6.51        & 100\%            & 4.71$\pm$4.55        & 86\%             \\
0.6                  & 4.98$\pm$6.49        & 97\%             & 2.10$\pm$2.86        & 58\%             \\ \hline
\multicolumn{1}{l}{} & \multicolumn{2}{c}{GraphEBM}            & \multicolumn{2}{c}{CDGS}                \\
$\delta$             & \textbf{Improvement} & \textbf{Success} & \textbf{Improvement} & \textbf{Success} \\ \hline
0.0                  & 15.75$\pm$7.40       & 99\%             & 12.83$\pm$7.01       & 100\%            \\
0.2                  & 8.40$\pm$6.38        & 94\%             & 11.70$\pm$6.84       & 100\%            \\
0.4                  & 4.95$\pm$5.90        & 79\%             & 9.56$\pm$6.33        & 100\%            \\
0.6                  & 3.15$\pm$5.08        & 45\%             & 5.10$\pm$5.80        & 98\%             \\ \hline
\end{tabular}%
}
\end{subfigure}
% \quad
\hfill
\begin{subfigure}[c]{0.49\textwidth}
\centering
\includegraphics[width=\textwidth]{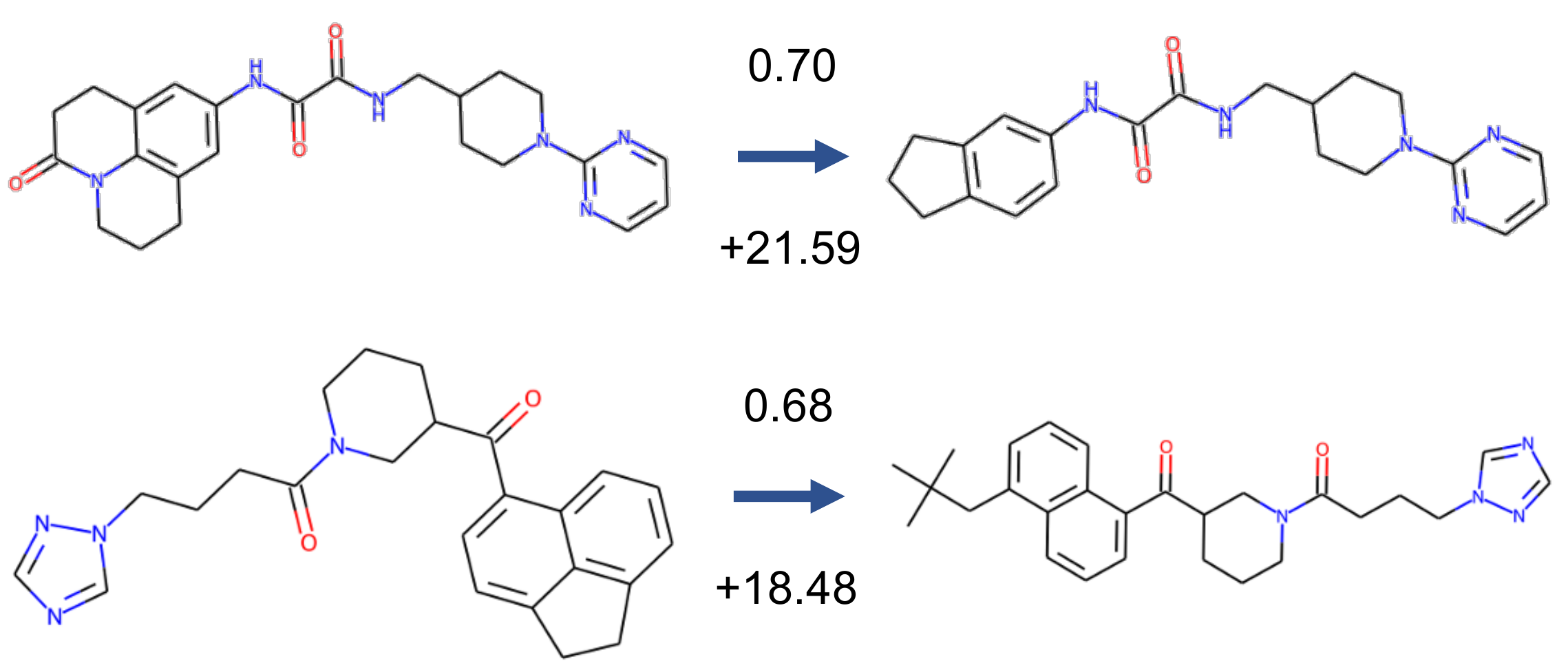}
\end{subfigure}
\end{table}

\begin{table*}[t]
\caption{Generation performance on generic graph datasets. The better results are indicated by a closer value with the performance of training graphs, and the best results are in bold.}
\label{tab:generic graph}
\centering
\renewcommand\arraystretch{1.3}
\resizebox{\textwidth}{!}{%
\begin{tabular}{ccccccccccccccccccccc}
\hline
              &  & \multicolumn{4}{c}{Community-small} &  & \multicolumn{4}{c}{Ego-small} &  & \multicolumn{4}{c}{Enzymes}  &  & \multicolumn{4}{c}{Ego}      \\ \cline{3-6} \cline{8-11} \cline{13-16} \cline{18-21} 
 &
   &
  \multicolumn{4}{c}{$|V|_{max}=20, |E|_{max}=62$} &
   &
  \multicolumn{4}{c}{$|V|_{max}=17, |E|_{max}=66$} &
   &
  \multicolumn{4}{c}{$|V|_{max}=125, |E|_{max}=149$} &
   &
  \multicolumn{4}{c}{$|V|_{max}=399, |E|_{max}=1071$} \\
 &
   &
  \multicolumn{4}{c}{$|V|_{avg}\approx15, |E|_{avg}\approx36$} &
   &
  \multicolumn{4}{c}{$|V|_{avg}\approx6, |E|_{avg}\approx9$} &
   &
  \multicolumn{4}{c}{$|V|_{avg}\approx33, |E|_{avg}\approx63$} &
   &
  \multicolumn{4}{c}{$|V|_{avg}\approx145, |E|_{avg}\approx335$} \\ \cline{3-6} \cline{8-11} \cline{13-16} \cline{18-21} 
              &  & Deg.    & Clus.   & Spec.   & GIN.  &  & Deg.   & Clus. & Spec. & GIN. &  & Deg.  & Clus. & Spec. & GIN. &  & Deg.  & Clus. & Spec. & GIN. \\ \hline
\textit{Train} &
   &
  \textit{0.035} &
  \textit{0.067} &
  \textit{0.045} &
  \textit{0.037} &
   &
  \textit{0.025} &
  \textit{0.029} &
  \textit{0.027} &
  \textit{0.016} &
   &
  \textit{0.011} &
  \textit{0.011} &
  \textit{0.011} &
  \textit{0.007} &
   &
  \textit{0.009} &
  \textit{0.009} &
  \textit{0.009} &
  \textit{0.005} \\ \hline
ER            &  & 0.300   & 0.239   & 0.100  & 0.278  &  & 0.200 & 0.094 & 0.361 & 0.230 &  & 0.844 & 0.381 & 0.104 & 0.808 &  & 0.738 & 0.397 & 0.868 & 0.118 \\
VGAE          &  & 0.391   & 0.257   & 0.095  & 0.360  &  & 0.146 & 0.046 & 0.249 & 0.089 &  & 0.811 & 0.514 & 0.153 & 0.716 &  & 0.873 & 1.210 & 0.935 & 0.520 \\
GraphRNN      &  & 0.106   & 0.115   & 0.091  & 0.353  &  & 0.155 & 0.229 & 0.167 & 0.472 &  & 0.397 & 0.302 & 0.260 & 1.495 &  & 0.140 & 0.755 & 0.316 & 1.283 \\
GraphRNN-U    &  & 0.410   & 0.297   & 0.103  & 0.970  &  & 0.471 & 0.416 & 0.398 & 0.915 &  & 0.932 & 1.000 & 0.367 & 1.263 &  & 1.413 & 1.097 & 1.110 & 1.317 \\
GRAN          &  & 0.125   & 0.164   & 0.111  & 0.196  &  & 0.096 & 0.072 & 0.095 & 0.106 &  & 0.215 & 0.147 & 0.034 & 0.069 &  & 0.594 & 0.425 & 1.025 & 0.244 \\
GRAN-U        &  & 0.106   & 0.127   & 0.083  & 0.164  &  & 0.155 & 0.229 & 0.167 & 0.094 &  & 0.343 & 0.122 & 0.041 & 0.242 &  & 0.099 & 0.170 & 0.179 & 0.128 \\
EDP-GNN       &  & 0.100   & 0.140   & 0.085  & 0.125  &  & 0.026 & 0.032 & 0.037 & 0.031 &  & 0.120 & 0.644 & 0.070 & 0.119 &  & 0.553 & 0.605 & 0.374 & 0.295 \\ 
GDSS          &  & 0.102   & 0.125   & 0.087  & 0.137  &  & 0.041 & 0.036 & 0.041 & 0.041 &  & 0.118 & 0.071 & 0.053 & 0.028 &  & 0.314 & 0.776 & 0.097 & 0.156 \\ \hline
CDGS-EM       &  & \textbf{0.052}   & \textbf{0.080}   & \textbf{0.064}  & \textbf{0.062}  &  & \textbf{0.025} & \textbf{0.031} & \textbf{0.033} & \textbf{0.025} &  & \textbf{0.048} & \textbf{0.070} & \textbf{0.033} & \textbf{0.024} &  & \textbf{0.036} & \textbf{0.075} & \textbf{0.026} & \textbf{0.026} \\
CDGS-GDPMS-30  &  & 0.100  & 0.121   & 0.084  & 0.120  &  & 0.116 & 0.064 & 0.141 & 0.052  &  & 0.140 & 0.127 & 0.041 & 0.040 &  & 0.157 & 0.109 & 0.153 & 0.064     \\ \hline
\end{tabular}%
}
\end{table*}

\subsubsection{Similarity-constrained Property Optimization}

We also show how our diffusion framework can be used for similarity-constrained property optimization.
Following \cite{Shigraphaf20, zang2020moflow}, we select $800$ molecules with low p-logP scores (\ie the octanol-water partition coefficients penalized by synthetic accessibility and number of long cycles) as the initial molecules for optimization.
We aim to generate new molecules with a higher p-logP while keeping similarity to the original molecules with a threshold $\delta$. 
The similarity metric is defined as Tanimoto similarity with Morgan fingerprints \cite{similarity_metric10}.
The property predictor is composed of $6$ hybrid message passing blocks with RGCN \cite{RGCN18} as the non-attention layer for differentiation.
We pretrain the time-dependent predictor on perturbed graphs of the ZINC250k dataset for $200$ epochs.
Each initial molecular graph is encoded into latent codes at the middle time $t_{\xi}=0.3$ through the forward-time ODE solver. After $50$ gradient ascent steps, all latent codes are decoded back to molecules with another gradient-guided reverse-time ODE solver. This procedure is repeated $20$ times with a different number of atoms to search for the highest property molecule that satisfies the similarity constraint.

Results for the similarity-constrained optimization are summarized in Table \ref{tab:const_opt}.
\textbf{GraphAF-RL} is the representative method combined with reinforcement learning, \textbf{MoFlow} is a flow-based method, and \textbf{GraphEBM} is an energy-based method for molecule optimization.
With the similarity constraint ($\delta > 0$), CDGS outperforms MoFlow and GraphEBM in terms of success rate and mean property improvement, showing competitive performance to the RL-based method.
Since RL-based methods require heavy property evaluator calls, which is unrealistic in some optimization scenarios, our framework could serve as a useful supplement for drug discovery tasks.
% Note that even though our method could explore the molecular space, training the predictor directly on the latent representations may provide more accurate guidance.

\subsection{Generic Graph Generation}
\subsubsection{Experimental Setup}
To display the graph structure distribution learning ability, we validate CDGS on four common generic graph datasets with various graph sizes and characteristics:
(1) \textit{Community-small}, $100$ two-community graphs generated by the Erd{\H{o}}s-R{\'e}nyi model (E-R) \cite{erdHos1960evolution} with $p=0.7$,
(2) \textit{Ego-small}, $200$ one-hop ego graphs extracted from Citeseer network \cite{sen2008collective}, 
(3) \textit{Enzymes}, $563$ protein graphs with more than $10$ nodes from BRENDA database \cite{schomburg2004brenda},
(4) \textit{Ego}, $757$ three-hop ego graphs extracted from Citeseer network \cite{sen2008collective}.
We use $8:2$ as the split ratio for train/test.
We generate $1024$ graphs for the evaluation on Community-small and Ego-small, and generate the same number of graphs as the test set on Enzymes and Ego. 
We follow the advice from \cite{o2021evaluation} to evaluate discrete graph structure distribution.
Three graph-level structure descriptor functions are selected: degree distribution (\textbf{Deg.}), clustering coefficient distribution (\textbf{Clus.}) and Laplacian spectrum histograms (\textbf{Spec.}).
We use MMD with the radial basis function kernel (RBF) to calculate the distance on features extracted by graph descriptors.
To accurately evaluate distribution distance, different from \cite{you2018graphrnn, Liao19GRAN, niu2020permutation} using a static smoothing hyperparameter for MMD, we provide a set of parameters and report the largest distance \cite{thompson2022Metric, huang2022graphgdp}. 
We also consider a well-established comprehensive neural-based metric (\textbf{GIN.}) from \cite{thompson2022Metric}.

\subsubsection{Baselines}
Apart from scored-based models (EDP-GNN and GDSS), we compare CDGS with a classical method (\textbf{ER} \cite{erdHos1960evolution}), a VAE-based method (\textbf{VGAE} \cite{kipf2016VGAE}), and two strong autoregressive graph generative models (\textbf{GraphRNN} \cite{you2018graphrnn}, \textbf{GRAN} \cite{Liao19GRAN}). \textbf{GraphRNN-U} and \textbf{GRAN-U} are trained with uniform node orderings to alleviate the bias from specific ordering strategies.

\subsubsection{Sampling Quality}
Table \ref{tab:generic graph} displays that, among four datasets, CDGS consistently achieves better performance than score-based models and autoregressive models.
Especially for the large Ego dataset, CDGS still generates graphs with high fidelity while the diffusion-based GDSS fails in Deg. and Clus. metrics.
The GDPMS is also supported for quick graph structure generation with acceptable quality. 
Thanks to the appropriate framework design and the emphasis on evolving discrete graph structures during the generative process, CDGS effectively captures the underlying distribution of graph topology.

\section{Conclusion}

We present a novel conditional diffusion model for molecular graph generation that takes advantage of discrete graph structure conditioning and delicate graph noise prediction model design.
Our model outperforms existing molecular graph generative methods in both graph space and chemical space for distribution learning, and also performs well for generic graph generation.
By adapting fast ODE solvers for graphs, we utilize our framework to make advances in efficient graph sampling and facilitate similarity-constrained optimization.
In the future, we plan to apply our model to molecule generation with complex conditions, such as target protein pockets.

\section*{Acknowledgment}
This work was supported in part by the National Natural Science Foundation of China (62272023, 51991395 and U21A20516).

% \section*{References}
% \medskip

\bibliography{sbmw}

%%%%%%%%%%%%%%%%%%%%%%%%%%%%%%%%%%%%%%%%%%%%%%%%%%%%%%%%%%%%

%%%%%%%%%%%%%%%%%%%%%%%%%%%%%%%%%%%%%%%%%%%%%%%%%%%%%%%%%%%%

\clearpage
\appendix

\section{Experimental Details}
\subsection{Hyperparameters}
The hyperparameters used for our CDGS in the experiments are provided in Table \ref{tab:hyper}.
In particular, we set the SDE setting to the default parameters of Variance Preserving SDE (VPSDE) without further sweeping, keeping the small signal-to-noise ratio at $\bm{G_T}$.
Different from GDSS \cite{JoLH22GDSS}, we adopt the unified SDE setting for $\bm{X}$ and $\bm{A}$ and utilize the simple EM solver, avoiding complex hyperparameter tuning.

\subsection{Molecular Graph Generation}
The dataset information is summarized in Table \ref{tab:mol_dataset}.

\begin{table}[!htbp]
\centering
\caption{Molecule dataset information.}
\label{tab:mol_dataset}
\renewcommand\arraystretch{1.4}
\resizebox{\columnwidth}{!}{%
\begin{tabular}{ccccc}
\hline
Dataset  & Number of molecules & Number of nodes  & Number of node types & Number of edge types \\ \hline
ZINC250k & 249,455             & $6\leq|V|\leq38$ & 9                    & 3                  \\
QM9      & 133,885             & $1\leq|V|\leq9$  & 4                    & 3                   \\ \hline
\end{tabular}%
}
\end{table}

\begin{table*}[!htbp]
\caption{Hyperparameters of CDGS used in graph generation experiments.}
\label{tab:hyper}
\centering
\renewcommand\arraystretch{1.5}
\resizebox{\textwidth}{!}{%
\begin{tabular}{cccccccc}
\hline
                       & Hyperparameter                   & ZINC250k     & QM9          & Community-small & Ego-small & Enzymes & Ego    \\ \hline
Data & Edge initial scale & $[-1.0,1.0]$ & $[-1.0,1.0]$ & $[-1.0,1.0]$ & $[-1.0,1.0]$ & $[-1.0,1.0]$ & $[-1.0,1.0]$ \\
                       & Node initial scale               & $[-0.5,0.5]$ & $[-0.5,0.5]$ & -               & -         & -       & -      \\ \hline
$\bm{\epsilon_\theta}$ & Number of message passing blocks & 10           & 6            & 6               & 3         & 6       & 3      \\
                       & Hidden dimension                 & 256          & 64           & 64              & 64        & 64      & 64     \\
                       & Number of attention heads        & 8            & 8            & 8               & 8         & 8       & 8      \\
                       & Number of Random Walks           & 20           & 8            & 16              & 8         & 24      & 20     \\ \hline
SDE                    & Type                             & VP           & VP           & VP              & VP        & VP      & VP     \\
                       & Number of EM sampling steps      & 1000         & 1000         & 1000            & 1000      & 1000    & 1000   \\
                       & $\beta_{min}$                    & 0.1          & 0.1          & 0.1             & 0.1       & 0.1     & 0.1    \\
                       & $\beta_{max}$                    & 20.0         & 20.0         & 20.0            & 20.0      & 20.0    & 20.0   \\ \hline
Train                  & Optimizer                        & Adam         & Adam         & Adam            & Adam      & Adam    & Adam   \\
                       & Learning rate                    & 1e-4         & 1e-4         & 1e-4            & 1e-4      & 1e-4    & 2e-4   \\
                       & Batch size                       & 64           & 128          & 64              & 64        & 48      & 8      \\
                       & Number of training steps         & 1.25M        & 1.0M         & 1.0M            & 0.8M      & 1.0M    & 0.8M   \\
                       & EMA                              & 0.9999       & 0.9999       & 0.9999          & 0.9999    & 0.9999  & 0.9999 \\ \hline
\end{tabular}%
}
\end{table*}

\subsubsection{Implementation Details}
For each molecule, we represent it with one-hot atom types $\{0, 1\}^{N \times F}$, ordinal edge types $\{0,1,2,3\}^{N \times N}$ (\ie no bonds, single bond, double bond and triple bond) and edge existence $\{0, 1\}^{N \times N}$.
We convert these variables to real numbers and obtain $\bm{G} = (\bm{X}, \bm{A})$.
Scaling and shifting are also used to adjust the initial number scale, making them simpler for neural networks to process.
As our method focuses on undirected graphs, we keep the adding noise and the output of edges symmetrical. 
We first sample the number of atoms from the probability mass function on the training graphs' atom number before the reverse generative process. 
After sampling through numerical solvers, we first move and shift the matrices back to their original scale and make quantization to obtain graph samples. 
We remain the biggest connected-subgraphs for those molecular graphs that are disconnected. 
The valency correction procedure from \cite{zang2020moflow} are adopted to further ensure molecular validity.
As for baselines, we report the performance from \cite{JoLH22GDSS}, and re-sample or retrain GDSS with its official code. 

\subsection{Generic Graph Generation}
\subsubsection{Implementation Details}
We directly use adjacency matrices $\{0,1\}^{N \times N}$ to represent generic graphs.
We still convert variables to real numbers and adjust their scale. 
For the MMD metrics (Deg., Clus., and Spec.) used in graph structure distribution evaluation, we choose a efficient positive definite kernel function, \ie an RBF kernel with a smoothing parameter $\upsilon$ denoted as 
\begin{equation}
    k(x_i, x_j) = \mathrm{exp}(\frac{-||x_i - x_j||^2}{2\upsilon^2}).
\end{equation}
It is important to choose $\upsilon$ to accurately measure the distribution distance.
We report the largest MMD values using a set of $\upsilon$ parameters.
$50$ candidate $\log\upsilon$ values are selected evenly between $[10^{-5}, 10^{5}]$.
We take $100$ bins for the histogram conversion of clustering coefficient and $200$ bins for the conversion of Laplacian spectrum.

As for the baselines, ER \cite{erdHos1960evolution} is implemented by the edge probability estimated by maximum likelihood on training graphs.
VGAE \cite{kipf2016VGAE} is a variational auto-encoder implemented by a graph convolution network encoder and a simple MLP decoder with inner product computation for edge existence.
For GraphRNN \cite{you2018graphrnn}, GRAN \cite{Liao19GRAN}, and
EDPGNN \cite{niu2020permutation}, we utilize their official code to train the models with the same data split and generate graphs for evaluation.

\subsection{Computing Resources}
We implement CDGS with PyTorch for all experiments and train the models on a single RTX A5000 GPU with AMD EPYC 7371 16-Core Processor.
The wall-clock times of different models are reported in the same environment.

\section{Algorithms}
We show the optimizing procedure in Algorithm \ref{alg:training} and the EM sampling procedure in Algorithm \ref{alg:em-solver}.
Moreover, we provide the implementation details of fast ODE solvers of different orders for in Algorithm \ref{alg:gdpms-1}, \ref{alg:gdpms-2}, \ref{alg:gdpms-3}, mainly derived from \cite{DPMS22}.
The solvers can be equipped with the gradient guidance from time-dependent molecule property predictor conveniently like Algorithm \ref{alg:gdpms-1-guide}. 

\algrenewcommand\algorithmicfunction{\textbf{def}}
% training
\begin{algorithm}[!htbp]
\caption{Optimizing CDGS}
\label{alg:training}
\textbf{Require}: original graph data $\bm{G_0}=(\bm{X_0},\bm{A_0})$, graph noise prediction model $\bm{\epsilon_\theta}$, schedule function $\alpha(\cdot)$ and $\sigma(\cdot)$, quantized function $quantize(\cdot)$

\begin{algorithmic}[1]
    \State Sample $t \sim \mathcal{U}(0,1], \bm{\epsilon}_{\bm{X}} \sim \mathcal{N}(\bm{0}, \bm{I}), \bm{\epsilon}_{\bm{A}} \sim \mathcal{N}(\bm{0}, \bm{I})$
    \State $\bm{G}_t = (\bm{X}_t,\bm{A}_t)  \gets (\alpha(t) \bm{X}_0 + \sigma(t) \bm{\epsilon}_{\bm{X}}, \alpha(t) \bm{A}_0 + \sigma(t) \bm{\epsilon}_{\bm{A}})$
    \State $\bar{\bm{A}}_t \gets quantize(\bm{A}_{t})$
    \State $\bm{\epsilon}_{\bm{\theta}}^{\bm{X}}, \bm{\epsilon}_{\bm{\theta}}^ {\bm{A}} \gets \bm{\epsilon}_{\bm{\theta}}(\bm{G}_t, \bar{\bm{A}_t}, t)$
    \State Minimize $||\bm{\epsilon}_{\bm{\theta}}^{\bm{X}} - \bm{\epsilon}_{\bm{X}}||^2_2 + 
    ||\bm{\epsilon}_{\bm{\theta}}^ {\bm{A}} - \bm{\epsilon}_{\bm{A}}||^2_2$
\end{algorithmic}
\end{algorithm}

% sampling with EM solvers
\begin{algorithm}[!htbp]
\caption{Sampling from CDGS with the Euler-Maruyama method}
\label{alg:em-solver}
\textbf{Require}: number of time steps $N$, graph noise prediction model $\bm{\epsilon_\theta}$, drift coefficient function $f(\cdot)$, diffusion coefficient function $g(\cdot)$, schedule function $\sigma(\cdot)$, quantized function $quantize(\cdot)$, post-processing function $post(\cdot)$

\begin{algorithmic}[1]
    \State Sample initial graph $\bm{G} \gets (\bm{X} \sim \mathcal{N}(\bm{0}, \bm{I}), \bm{A} \sim \mathcal{N}(\bm{0}, \bm{I}) )$, 
    \State $\Delta t = \frac{T}{N}$
    \For{$i \gets N$ to $1$}
    \State $\bar{\bm{A}} \gets quantize(\bm{A})$
    \State $\bm{\epsilon}_{\bm{X}} \sim \mathcal{N}(\bm{0}, \bm{I}), \bm{\epsilon}_{\bm{A}} \sim \mathcal{N}(\bm{0}, \bm{I})$
    \State $t \gets i\Delta t$
    \State $\bm{\epsilon}_{\bm{\theta}}^{\bm{X}}, \bm{\epsilon}_{\bm{\theta}}^ {\bm{A}} \gets \bm{\epsilon}_{\bm{\theta}}(\bm{G}, \bar{\bm{A}}, t)$
    \State $\bm{X} \gets \bm{X} - (f(t)\bm{X}+\frac{g(t)^2}{\sigma(t)}\bm{\epsilon}_{\bm{\theta}}^{\bm{X}})\Delta t + g(t)\sqrt{\Delta t} \bm{\epsilon}_{\bm{X}}$
    \State $\bm{A} \gets \bm{A} - (f(t)\bm{A}+\frac{g(t)^2}{\sigma(t)}\bm{\epsilon}_{\bm{\theta}}^{\bm{A}})\Delta t + g(t)\sqrt{\Delta t} \bm{\epsilon}_{\bm{A}}$
    \EndFor
    \State \Return $post(\bm{X}, \bm{A})$
\end{algorithmic}
\end{algorithm}

\begin{algorithm}[!htbp]
\caption{Graph DPM-Solver 1}
\label{alg:gdpms-1}
\textbf{Require}: initial graph $\bm{G_T}=(\bm{X_T},\bm{A_T})$, time step schedule $\{t_i\}_{i=0}^M$, graph noise prediction model $\bm{\epsilon_\theta}$, quantized function $quantize(\cdot)$, post-processing function $post(\cdot)$

\begin{algorithmic}[1]
    \Function{gdpms-1}{$\bm{\tilde{X}}_{t_{i-1}}, \bm{\tilde{A}}_{t_{i-1}}, t_{i-1}, t_i$}
        \State $h_i \gets \lambda_{t_i} - \lambda_{t_{i-1}}$
        \State $\bm{\bar{A}}_{t_{i-1}}' \gets quantize(\bm{\tilde{A}}_{t_{i-1}})$
        \State $\bm{\tilde{\epsilon}}_{t_{i-1}}^{\bm{X}}, \  \bm{\tilde{\epsilon}}_{t_{i-1}}^{\bm{A}} \gets \bm{\epsilon_{\theta}}((\bm{\tilde{X}}_{t_{i-1}}, \bm{\tilde{A}}_{t_{i-1}}), {\bm{\bar{A}}}_{t_{i-1}}', t_{i-1})$
        \State $\bm{\tilde{X}}_{t_i} \gets \frac{\alpha_{t_i}}{\alpha_{t_{i-1}}} \bm{\tilde{X}}_{t_{i-1}} - 
        \sigma_{t_i}(e^{h_i} - 1) \bm{\tilde{\epsilon}}_{t_{i-1}}^{\bm{X}} $
        \State $\bm{\tilde{A}}_{t_i} \gets \frac{\alpha_{t_i}}{\alpha_{t_{i-1}}} \bm{\tilde{A}}_{t_{i-1}} - 
        \sigma_{t_i}(e^{h_i} - 1) \bm{\tilde{\epsilon}}_{t_{i-1}}^{\bm{A}}$
        \State \Return $\bm{\tilde{X}}_{t_i}, \bm{\tilde{A}}_{t_i}$ 
    \EndFunction
    \State $\bm{\tilde{X}}_{t_0}, \bm{\tilde{A}}_{t_0} \gets \bm{X_T},\bm{A_T}$
    \For{$i \gets 1$ to $M$}
        \State $\bm{\tilde{X}}_{t_i}, \bm{\tilde{A}}_{t_i} \gets
            $ GDPMS-1$(\bm{\tilde{X}}_{t_{i-1}}, \bm{\tilde{A}}_{t_{i-1}}, t_{i-1}, t_i)$
    \EndFor
    \State \Return $post(\bm{\tilde{X}}_{t_M}, \bm{\tilde{A}}_{t_M})$
\end{algorithmic}
\end{algorithm}

\begin{algorithm*}[!htbp]
\caption{Graph DPM-Solver 2}
\label{alg:gdpms-2}
\textbf{Require}: initial graph $\bm{G_T}=(\bm{X_T},\bm{A_T})$, time step schedule $\{t_i\}_{i=0}^M$, graph noise prediction model $\bm{\epsilon_\theta}$, quantized function $quantize(\cdot)$, post-processing function $post(\cdot)$, $r_1=0.5$

\begin{algorithmic}[1]
    \Function{gdpms-2}{$\bm{\tilde{X}}_{t_{i-1}}, \bm{\tilde{A}}_{t_{i-1}}, t_{i-1}, t_i, r_1$}
        \State $h_i \gets \lambda_{t_i} - \lambda_{t_{i-1}}$
        \State $s_i \gets t_\lambda (\lambda_{t_{i-1}} + r_1h_i)$
        \State $\bm{\bar{A}}_{t_{i-1}}' \gets quantize(\bm{\tilde{A}}_{t_{i-1}})$
        
        \State $\bm{\tilde{\epsilon}}_{t_{i-1}}^{\bm{X}}, \  \bm{\tilde{\epsilon}}_{t_{i-1}}^{\bm{A}} \gets \bm{\epsilon_{\theta}}((\bm{\tilde{X}}_{t_{i-1}}, \bm{\tilde{A}}_{t_{i-1}}), {\bm{\bar{A}}}_{t_{i-1}}', t_{i-1})$
        
        \State $\bm{u}_{i}^{\bm{X}} \gets \frac{\alpha_{s_i}}{\alpha_{t_{i-1}}} \bm{\tilde{X}}_{t_{i-1}} - \sigma_{s_i}(e^{r_1h_i} - 1) \bm{\tilde{\epsilon}}_{t_{i-1}}^{\bm{X}}$
        
        \State $\bm{u}_{i}^{\bm{A}} \gets \frac{\alpha_{s_i}}{\alpha_{t_{i-1}}} \bm{\tilde{A}}_{t_{i-1}} - \sigma_{s_i}(e^{r_1h_i} - 1) \bm{\tilde{\epsilon}}_{t_{i-1}}^{\bm{A}}$
        
        \State $\bm{u}_{i}^{\bm{\bar{A}}} \gets quantize(\bm{u}_{i}^{\bm{A}})$
        
        \State $\bm{\tilde{\epsilon}}_{s_i}^{\bm{X}}, \  \bm{\tilde{\epsilon}}_{s_i}^{\bm{A}} \gets \bm{\epsilon_{\theta}}((\bm{u}_{i}^{\bm{X}}, \bm{u}_{i}^{\bm{A}}), \bm{u}_{i}^{\bm{\bar{A}}}, s_{i})$
        
        \State $\bm{\tilde{X}}_{t_i} \gets \frac{\alpha_{t_i}}{\alpha_{t_{i-1}}}\bm{\tilde{X}}_{t_{i-1}} 
        - \sigma_{t_i}(e^{h_i} - 1) \bm{\tilde{\epsilon}}_{t_{i-1}}^{\bm{X}}
        - \frac{\sigma_{t_i}}{2r_i}(e^{h_i} - 1) (\bm{\tilde{\epsilon}}_{s_i}^{\bm{X}} - \bm{\tilde{\epsilon}}_{t_{i-1}}^{\bm{X}}) $
        
        \State $\bm{\tilde{A}}_{t_i} \gets \frac{\alpha_{t_i}}{\alpha_{t_{i-1}}}\bm{\tilde{A}}_{t_{i-1}} 
        - \sigma_{t_i}(e^{h_i} - 1) \bm{\tilde{\epsilon}}_{t_{i-1}}^{\bm{A}}
        - \frac{\sigma_{t_i}}{2r_i}(e^{h_i} - 1) (\bm{\tilde{\epsilon}}_{s_i}^{\bm{A}} - \bm{\tilde{\epsilon}}_{t_{i-1}}^{\bm{A}}) $
        
        \State \Return $\bm{\tilde{X}}_{t_i}, \bm{\tilde{A}}_{t_i}$ 
    \EndFunction
    \State $\bm{\tilde{X}}_{t_0}, \bm{\tilde{A}}_{t_0} \gets \bm{X_T},\bm{A_T}$
    \For{$i \gets 1$ to $M$}
        \State $\bm{\tilde{X}}_{t_i}, \bm{\tilde{A}}_{t_i} \gets
            $ GDPMS-2$(\bm{\tilde{X}}_{t_{i-1}}, \bm{\tilde{A}}_{t_{i-1}}, t_{i-1}, t_i, r_1)$
    \EndFor
    \State \Return $post(\bm{\tilde{X}}_{t_M}, \bm{\tilde{A}}_{t_M})$
\end{algorithmic}
\end{algorithm*}

\begin{algorithm*}[!htbp]
\caption{Graph DPM-Solver 3}
\label{alg:gdpms-3}
\textbf{Require}: initial graph $\bm{G_T}=(\bm{X_T},\bm{A_T})$, time step schedule $\{t_i\}_{i=0}^M$, graph noise prediction model $\bm{\epsilon_\theta}$, quantized function $quantize(\cdot)$, post-processing function $post(\cdot)$, $r_1=\frac{1}{3}$, $r_2=\frac{2}{3}$

\begin{algorithmic}[1]
    \Function{gdpms-3}{$\bm{\tilde{X}}_{t_{i-1}}, \bm{\tilde{A}}_{t_{i-1}}, t_{i-1}, t_i, r_1, r_2$}
        \State $h_i \gets \lambda_{t_i} - \lambda_{t_{i-1}}$
        \State $s_{2i-1} \gets t_\lambda (\lambda_{t_{i-1}} + r_1h_i), \quad s_{2i} \gets t_\lambda (\lambda_{t_{i-1}} + r_2h_i)$
        
        \State $\bm{\bar{A}}_{t_{i-1}}' \gets quantize(\bm{\tilde{A}}_{t_{i-1}})$
        
        \State $\bm{\tilde{\epsilon}}_{t_{i-1}}^{\bm{X}}, \  \bm{\tilde{\epsilon}}_{t_{i-1}}^{\bm{A}} \gets \bm{\epsilon_{\theta}}((\bm{\tilde{X}}_{t_{i-1}}, \bm{\tilde{A}}_{t_{i-1}}), {\bm{\bar{A}}}_{t_{i-1}}', t_{i-1})$
        
        \State $\bm{u}_{2i-1}^{\bm{X}} \gets \frac{\alpha_{s_{2i-1}}}{\alpha_{t_{i-1}}} \bm{\tilde{X}}_{t_{i-1}} - \sigma_{s_{2i-1}}(e^{r_1h_i} - 1) \bm{\tilde{\epsilon}}_{t_{i-1}}^{\bm{X}}$
        
        \State $\bm{u}_{2i-1}^{\bm{A}} \gets \frac{\alpha_{s_{2i-1}}}{\alpha_{t_{i-1}}} \bm{\tilde{A}}_{t_{i-1}} - \sigma_{s_{2i-1}}(e^{r_1h_i} - 1) \bm{\tilde{\epsilon}}_{t_{i-1}}^{\bm{A}}$
        
        \State $\bm{u}_{2i-1}^{\bm{\bar{A}}} \gets quantize(\bm{u}_{2i-1}^{\bm{A}})$
        
        \State $\bm{\tilde{\epsilon}}_{s_{2i-1}}^{\bm{X}}, \  \bm{\tilde{\epsilon}}_{s_{2i-1}}^{\bm{A}} \gets \bm{\epsilon_{\theta}}((\bm{u}_{2i-1}^{\bm{X}}, \bm{u}_{2i-1}^{\bm{A}}), \bm{u}_{2i-1}^{\bm{\bar{A}}}, s_{2i-1})$
        
        \State $\bm{D}_{2i-1}^{\bm{X}} \gets \bm{\tilde{\epsilon}}_{s_{2i-1}}^{\bm{X}} - \bm{\tilde{\epsilon}}_{t_{i-1}}^{\bm{X}}, \quad
        \bm{D}_{2i-1}^{\bm{A}} \gets \bm{\tilde{\epsilon}}_{s_{2i-1}}^{\bm{A}} - \bm{\tilde{\epsilon}}_{t_{i-1}}^{\bm{A}}$
        
        \State $\bm{u}_{2i}^{\bm{X}} \gets \frac{\alpha_{s_{2i}}}{\alpha_{t_{i-1}}} \bm{\tilde{X}}_{t_{i-1}} 
        - \sigma_{s_{2i}}(e^{r_2h_i} - 1) \bm{\tilde{\epsilon}}_{t_{i-1}}^{\bm{X}} 
        - \frac{\sigma_{s_{2i}}r_2}{r_1} (\frac{e^{r_2h_i}-1}{r_2h_i} - 1) \bm{D}_{2i-1}^{\bm{X}}$
        
        \State $\bm{u}_{2i}^{\bm{A}} \gets \frac{\alpha_{s_{2i}}}{\alpha_{t_{i-1}}} \bm{\tilde{A}}_{t_{i-1}} 
        - \sigma_{s_{2i}}(e^{r_2h_i} - 1) \bm{\tilde{\epsilon}}_{t_{i-1}}^{\bm{A}} 
        - \frac{\sigma_{s_{2i}}r_2}{r_1} (\frac{e^{r_2h_i}-1}{r_2h_i} - 1) \bm{D}_{2i-1}^{\bm{A}}$
        
        \State $\bm{u}_{2i}^{\bm{\bar{A}}} \gets quantize(\bm{u}_{2i}^{\bm{A}})$
        
        \State $\bm{\tilde{\epsilon}}_{s_{2i}}^{\bm{X}}, \  \bm{\tilde{\epsilon}}_{s_{2i}}^{\bm{A}} \gets \bm{\epsilon_{\theta}}((\bm{u}_{2i}^{\bm{X}}, \bm{u}_{2i}^{\bm{A}}), \bm{u}_{2i}^{\bm{\bar{A}}}, s_{2i})$
        
        \State $\bm{D}_{2i}^{\bm{X}} \gets \bm{\tilde{\epsilon}}_{s_{2i}}^{\bm{X}} - \bm{\tilde{\epsilon}}_{t_{i-1}}^{\bm{X}}, \quad
        \bm{D}_{2i}^{\bm{A}} \gets \bm{\tilde{\epsilon}}_{s_{2i}}^{\bm{A}} - \bm{\tilde{\epsilon}}_{t_{i-1}}^{\bm{A}}$
        
        \State $\bm{\tilde{X}}_{t_i} \gets \frac{\alpha_{t_i}}{\alpha_{t_{i-1}}}\bm{\tilde{X}}_{t_{i-1}} 
        - \sigma_{t_i}(e^{h_i} - 1) \bm{\tilde{\epsilon}}_{t_{i-1}}^{\bm{X}}
        - \frac{\sigma_{t_i}}{r_i}(\frac{e^{h_i} - 1}{h} - 1)  \bm{D}_{2i}^{\bm{X}}$
        
        \State $\bm{\tilde{A}}_{t_i} \gets \frac{\alpha_{t_i}}{\alpha_{t_{i-1}}}\bm{\tilde{A}}_{t_{i-1}} 
        - \sigma_{t_i}(e^{h_i} - 1) \bm{\tilde{\epsilon}}_{t_{i-1}}^{\bm{A}}
        - \frac{\sigma_{t_i}}{r_i}(\frac{e^{h_i} - 1}{h} - 1)  \bm{D}_{2i}^{\bm{A}}$
        
        \State \Return $\bm{\tilde{X}}_{t_i}, \bm{\tilde{A}}_{t_i}$ 
    \EndFunction
    \State $\bm{\tilde{X}}_{t_0}, \bm{\tilde{A}}_{t_0} \gets \bm{X_T},\bm{A_T}$
    \For{$i \gets 1$ to $M$}
        \State $\bm{\tilde{X}}_{t_i}, \bm{\tilde{A}}_{t_i} \gets
            $ GDPMS-3$(\bm{\tilde{X}}_{t_{i-1}}, \bm{\tilde{A}}_{t_{i-1}}, t_{i-1}, t_i, r_1, r_2)$
    \EndFor
    \State \Return $post(\bm{\tilde{X}}_{t_M}, \bm{\tilde{A}}_{t_M})$
\end{algorithmic}
\end{algorithm*}

\begin{algorithm*}[!htbp]
\caption{Graph DPM-Solver 1 with gradient guidance}
\label{alg:gdpms-1-guide}
\textbf{Require}: initial graph $\bm{G_T}=(\bm{X_T},\bm{A_T})$, time step schedule $\{t_i\}_{i=0}^M$, graph noise prediction model $\bm{\epsilon_\theta}$, quantized function $quantize(\cdot)$, post-processing function $post(\cdot)$, property predictor $\bm{R_{\psi}}$, guidance weight $r$

\begin{algorithmic}[1]
    \Function{gdpms-1-guide}{$\bm{\tilde{X}}_{t_{i-1}}, \bm{\tilde{A}}_{t_{i-1}}, t_{i-1}, t_i, r$}
        \State $h_i \gets \lambda_{t_i} - \lambda_{t_{i-1}}$
        \State $\bm{\bar{A}}_{t_{i-1}}' \gets quantize(\bm{\tilde{A}}_{t_{i-1}})$
        \State $\bm{\tilde{\epsilon}}_{t_{i-1}}^{\bm{X}}, \  \bm{\tilde{\epsilon}}_{t_{i-1}}^{\bm{A}} \gets \bm{\epsilon_{\theta}}((\bm{\tilde{X}}_{t_{i-1}}, \bm{\tilde{A}}_{t_{i-1}}), {\bm{\bar{A}}}_{t_{i-1}}', t_{i-1})$
        
        \State $\bm{R}_{t_{i-1}} = \bm{R_\psi}((\bm{\tilde{X}}_{t_{i-1}}, \bm{\tilde{A}}_{t_{i-1}}), t_{i-1})$
        
        \State $\bm{\tilde{X}}_{t_i} \gets \frac{\alpha_{t_i}}{\alpha_{t_{i-1}}} \bm{\tilde{X}}_{t_{i-1}} - 
        \sigma_{t_i}(e^{h_i} - 1) (\bm{\tilde{\epsilon}}_{t_{i-1}}^{\bm{X}} - r\sigma_{t_{i-1}} \nabla_{\bm{X}}^{*}\bm{R}_{t_{i-1}})$
        
        \State $\bm{\tilde{A}}_{t_i} \gets \frac{\alpha_{t_i}}{\alpha_{t_{i-1}}} \bm{\tilde{A}}_{t_{i-1}} - 
        \sigma_{t_i}(e^{h_i} - 1) (\bm{\tilde{\epsilon}}_{t_{i-1}}^{\bm{A}} - r\sigma_{t_{i-1}} \nabla_{\bm{A}}^{*}\bm{R}_{t_{i-1}}))$
        
        \State \Return $\bm{\tilde{X}}_{t_i}, \bm{\tilde{A}}_{t_i}$ 
    \EndFunction
    \State $\bm{\tilde{X}}_{t_0}, \bm{\tilde{A}}_{t_0} \gets \bm{X_T},\bm{A_T}$
    \For{$i \gets 1$ to $M$}
        \State $\bm{\tilde{X}}_{t_i}, \bm{\tilde{A}}_{t_i} \gets
            $ GDPMS-1-GUIDE$(\bm{\tilde{X}}_{t_{i-1}}, \bm{\tilde{A}}_{t_{i-1}}, t_{i-1}, t_i, r)$
    \EndFor
    \State \Return $post(\bm{\tilde{X}}_{t_M}, \bm{\tilde{A}}_{t_M})$
\end{algorithmic}
\end{algorithm*}

\section{Visualization}
\begin{figure*}[!htbp]
    \centering
    \includegraphics[width=\textwidth]{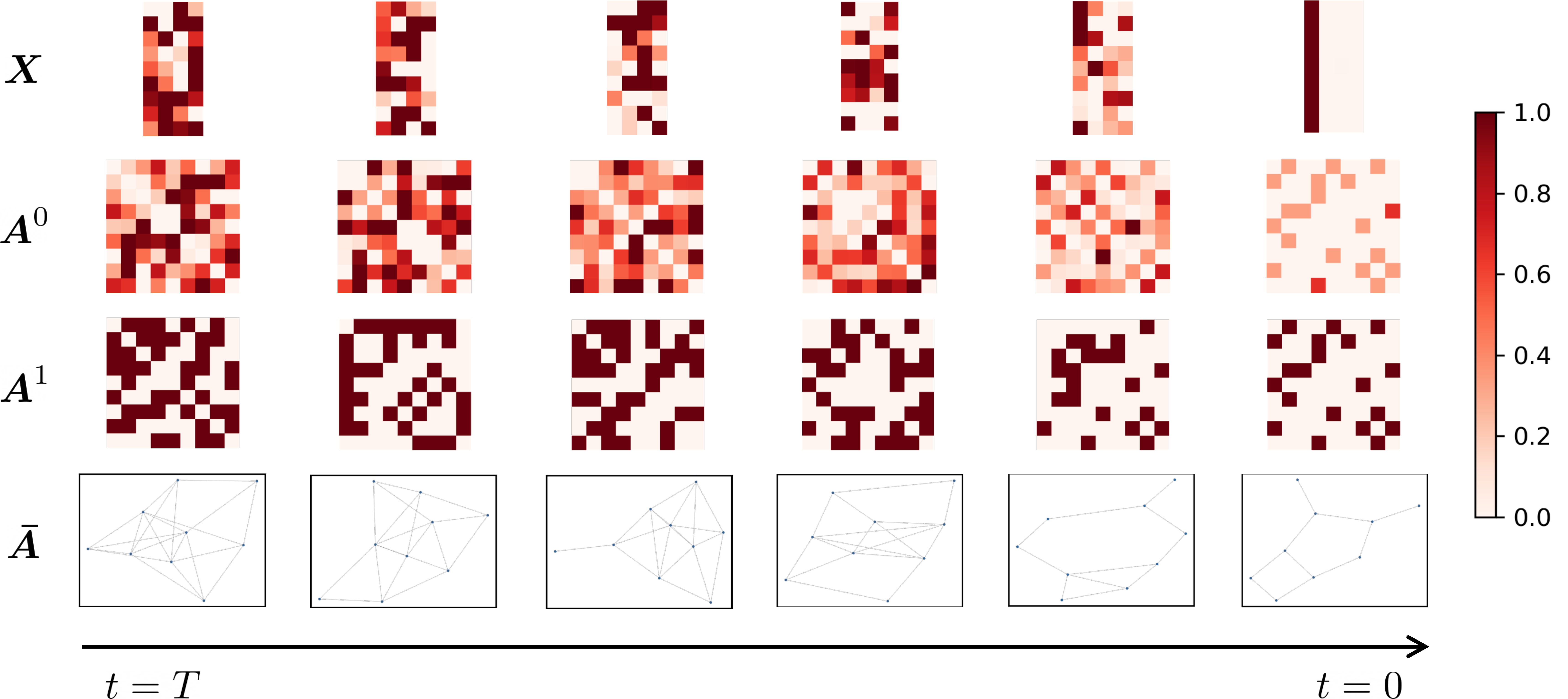}
    \caption{Molecular graph normalized visualization at different steps in the reverse generative process from a model trained on QM9. $\bm{X}$ is the node feature matrix, $\bm{A}^0$ is the edge type matrix, and $\bm{A}^1$ is the quantized edge existence matrix.}
    \label{fig:qm9_diff}
\end{figure*}

\begin{figure*}[!htbp]
    \centering
    \includegraphics[width=\textwidth]{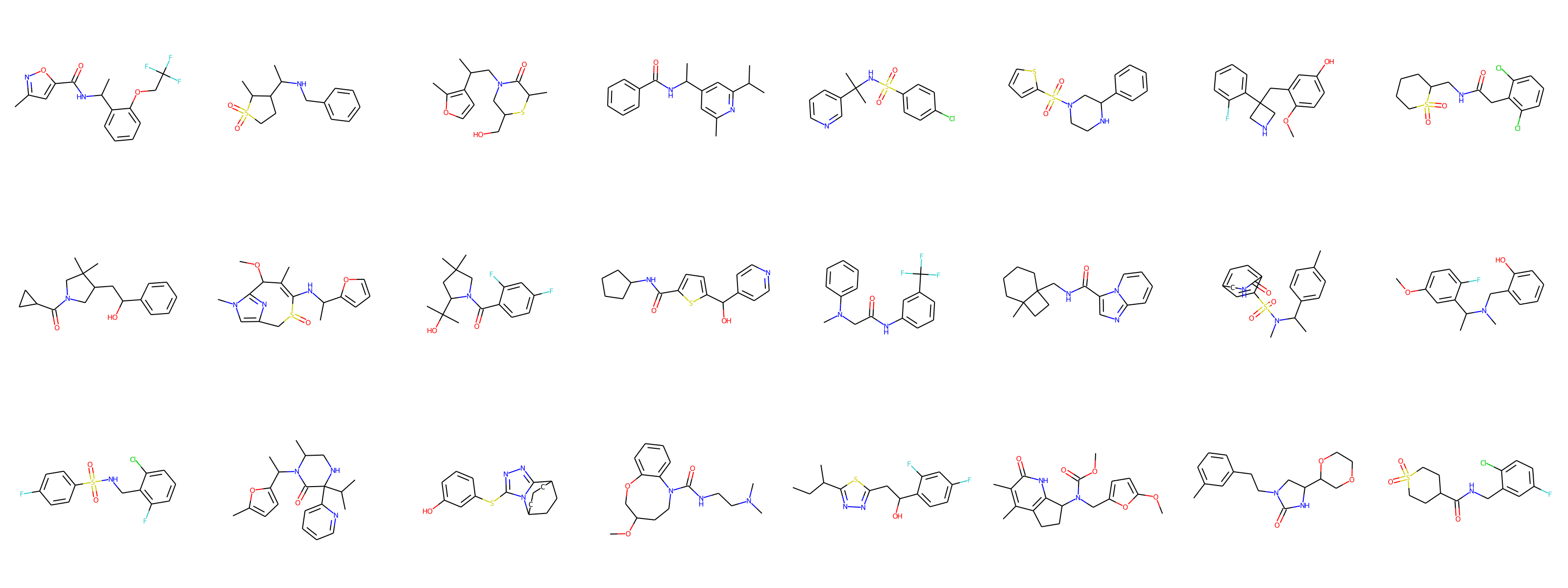}
    \caption{The generated samples from the model trained on the ZINC250k dataset.}
    \label{fig:zink_vis}
\end{figure*}

\begin{figure*}[!htbp]
    \centering
    \includegraphics[width=\textwidth]{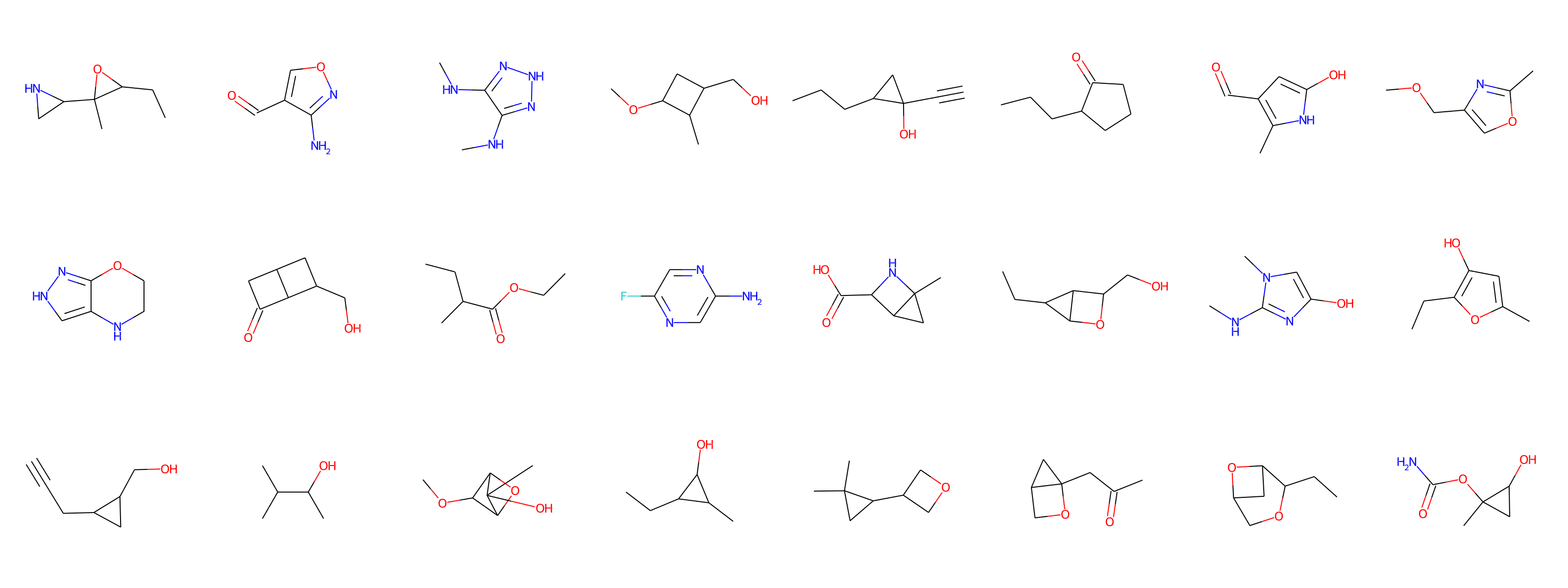}
    \caption{The generated samples from the model trained on the QM9 dataset.}
    \label{fig:qm9_vis}
\end{figure*}

\begin{figure*}[!htbp]
    \centering
    \includegraphics[width=0.75\textwidth]{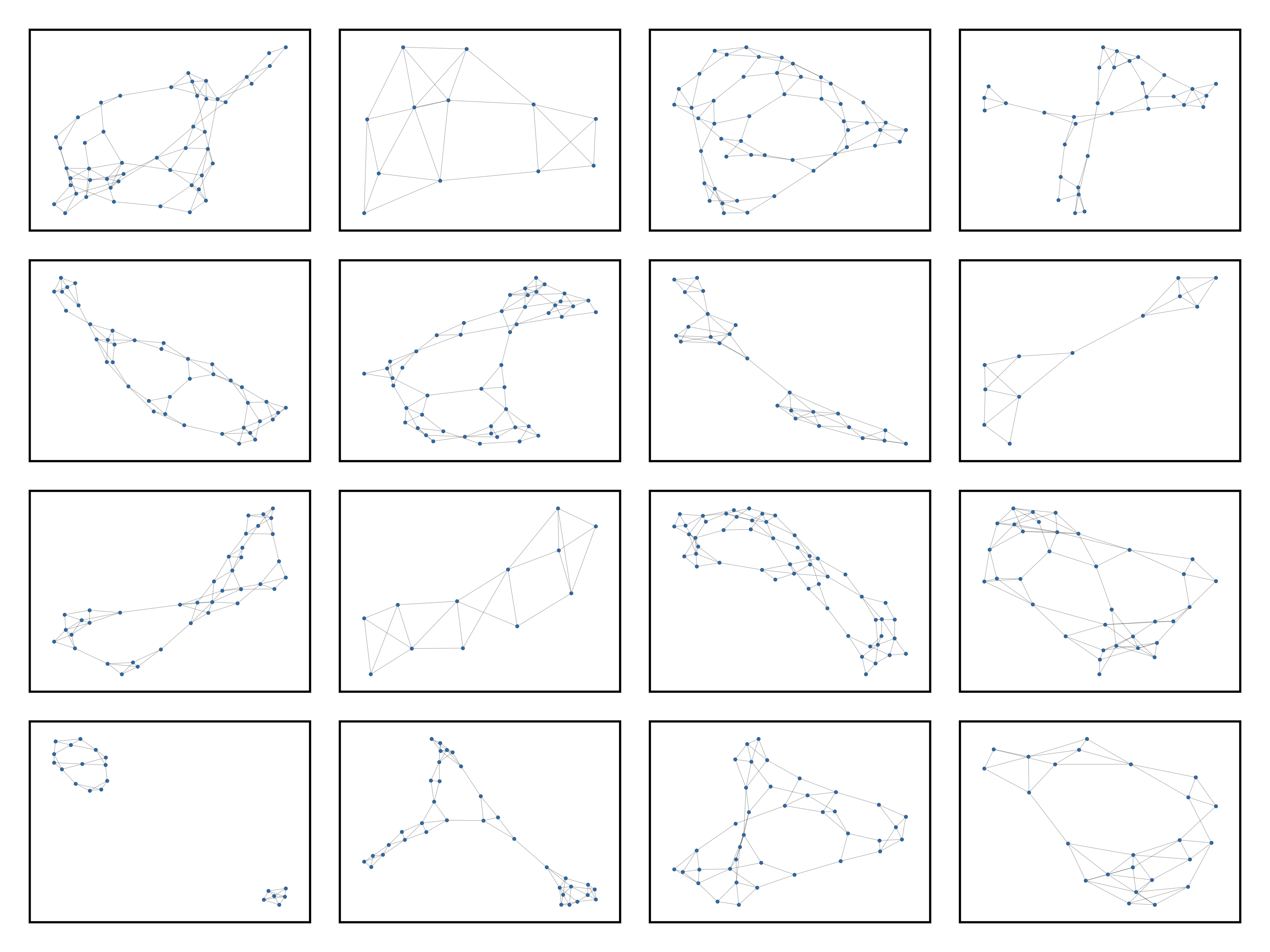}
    \caption{The generated samples from the model trained on the Enzymes dataset.}
    \label{fig:enzymes_vis}
\end{figure*}

\begin{figure*}[!htbp]
    \centering
    \includegraphics[width=0.75\textwidth]{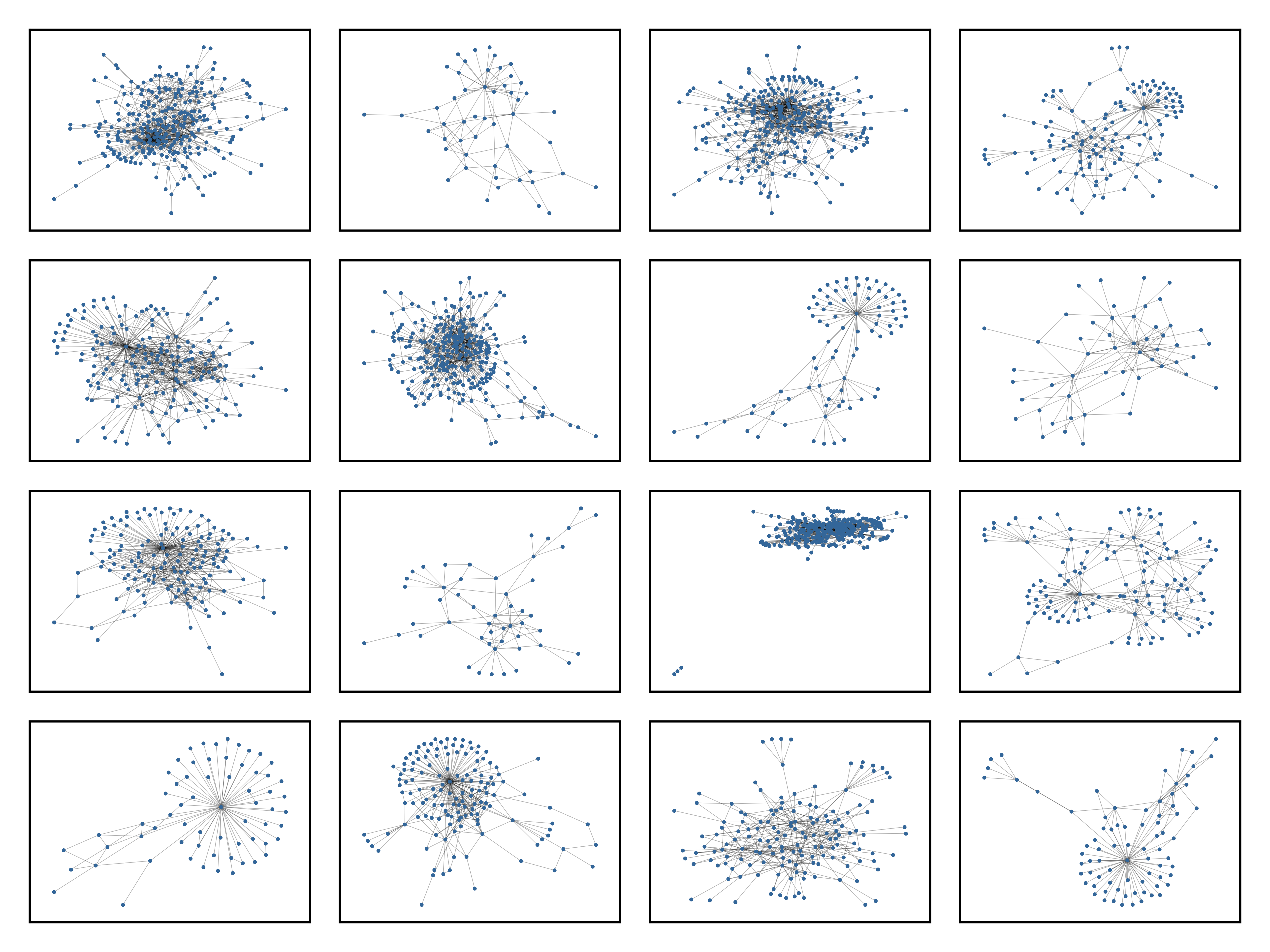}
    \caption{The generated samples from the model trained on the Ego dataset.}
    \label{fig:ego_vis}
\end{figure*}

\begin{figure*}[!htbp]
    \centering
    \includegraphics[width=0.75\textwidth]{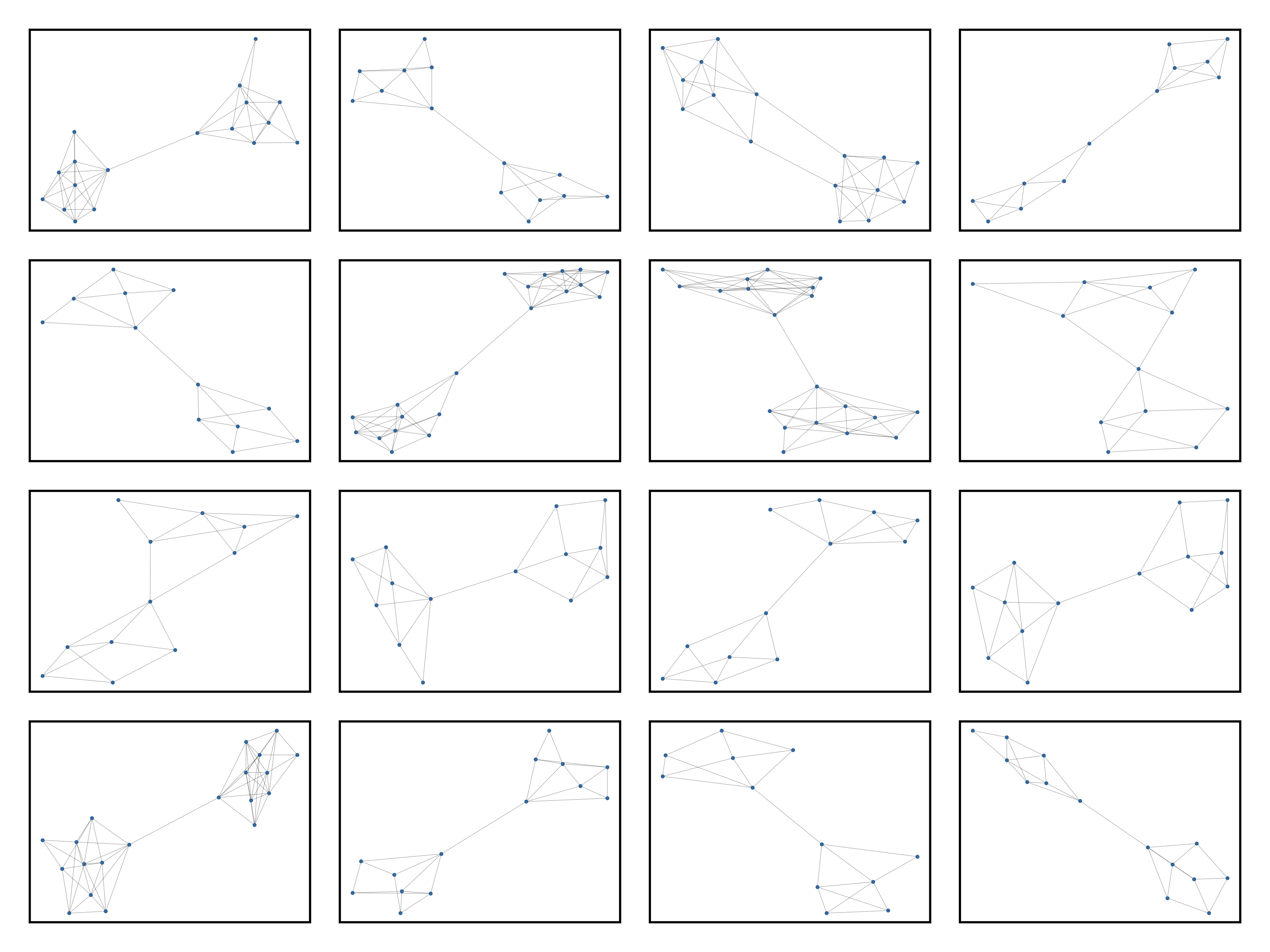}
    \caption{The generated samples from the model trained on the Community-small dataset.}
    \label{fig:coms_vis}
\end{figure*}

We visualize the reverse generative process on the QM9 dataset in Figure \ref{fig:qm9_diff}. 
We provides the visualization of generated graphs on different datasets: ZINC250k (in Figure \ref{fig:zink_vis}), QM9 (in Figure \ref{fig:qm9_vis}), Enzymes (in Figure \ref{fig:enzymes_vis}), Ego (in Figure \ref{fig:ego_vis}), and Community-small (in Figure \ref{fig:coms_vis}).

\end{document}